\documentclass[lettersize,journal]{IEEEtran}
\usepackage{amsmath,amsfonts}

\usepackage{caption}
\usepackage{mathtools}
\usepackage{nomencl}
\usepackage{ntheorem}   % for theorems
\usepackage{lipsum}     % for sample text
\usepackage{graphicx}
\usepackage{algpseudocode}

\usepackage{algorithm}
\usepackage{changes}
\usepackage{xpatch}
\usepackage{multicol}
\usepackage{graphics}
\usepackage{cite}
 \usepackage{amssymb}
\usepackage{cite}
\usepackage{xcolor}
\usepackage{bm}
\usepackage{amsmath}

\DeclareRobustCommand{\uvec}[1]{{%
		\ifcsname uvec#1\endcsname
		\csname uvec#1\endcsname
		\else
		\bm{\mathbf{#1}}%
		\fi
}}
\usepackage{algpseudocode}
\usepackage{algorithm}
\usepackage{algpascal}

\usepackage{float}
\usepackage{flushend}
\usepackage{hyperref}

\hypersetup{
    colorlinks=true,
    linkcolor=blue,
    filecolor=green, 
    urlcolor=black,
    citecolor=green
}
\usepackage{balance}

\begin{document}
\title{PSM: A Predictive Safety Model for Body Motion Based On the Spring-Damper Pendulum }

\author{Seyed Amir Tafrishi, Ankit A. Ravankar and Yasuhisa Hirata % <-this % Mstops a space
	\thanks{Authors are with Department of Robotics, Tohoku University, 6-
		6-01 Aramaki-Aoba, Aoba-ku, Sendai 980-8579, Japan. 
		{\tt\small \{s.a.tafrishi, ankit, hirata\}@srd.mech.tohoku.ac.jp}	}%
}

\maketitle

%%%%%%%%%%%%%%%%%%%%%%%%%%%%%%%%%%%%%%%%%%%%%%%%%%%%%%%%%%%%%%%%%%%%%%%%%%%%%%%%
\begin{abstract}
Quantifying the safety of the human body orientation is an important issue in human-robot interaction. Knowing the changing physical constraints on human motion can improve inspection of safe human motions and bring essential information about stability and normality of human body orientations with real-time risk assessment. Also, this information can be used in cooperative robots and monitoring systems to evaluate and interact in the environment more freely. Furthermore, the workspace area can be more deterministic with the known physical characteristics of safety. Based on this motivation, we propose a novel predictive safety model (PSM) that relies on the information of an inertial measurement unit on the human chest. The PSM encompasses a 3-Dofs spring-damper pendulum model that predicts human motion based on a safe motion dataset. The estimated safe orientation of humans is obtained by integrating a safety dataset and an elastic spring-damper model in a way that the proposed approach can realize complex motions at different safety levels. We did experiments in a real-world scenario to verify our novel proposed model. This novel approach can be used in different guidance/assistive robots and health monitoring systems to support and evaluate the human condition, particularly elders.
\end{abstract}

%%%%%%%%%%%%%%%%%%%%%%%%%%%%%%%%%%%%%%%%%%%%%%%%%%%%%%%%%%%%%%%%%%%%%%%%%%%%%%%%
\section{Introduction}
Human safety is an important topic in the field of robotics. Understating how safe human moves can bring different developments for guidance/health monitoring systems and assistive/cooperative robots. In this work, we define safety as the ability to know how human orientation/motion are normal (with distinguishing the unstable and unpredictable behaviours) based on a kinematic/dynamic reference. Additionally, this factor can determine how much robots or other systems can co-exist with humans in the same environment. The risk assessment is rather limited based on current ISO standards, e.g., ISO/TS 15066 \cite{iso2011iso} and ISO 10218 \cite{iso2016ts}  in robotics because the definition of human safety is quite challenging to quantify and requires fixed safeguards. This reason mainly arises from the complexity of both human motion and the lack of information about human intentions as awareness means. This problem prevents the robots from being highly active in places where human moves or interacts. 
%-------- Done

By involving automation and robotics into the industry in the early 1970s, the attention to the safety of humans has become important. Since then, there have been pioneering research \cite{bixby1991proactive,alami2006safe,colgate2008safety,vasic2013safety} in the definition of safety where humans and robots coexist. The safety problem can be looked at from two sides: humans and robots. The main concern from the robot's point-of-view was improving the automation and control techniques of different mechanisms e.g., manipulators to create a safe workspace. This motivated various strategies in robot manipulation with human collaboration in loop \cite{ikeura1995variable,li2017adaptive} and fixed safeguard. Then, a planning strategy for a manipulator was proposed with considering the psychological safety of the human collaborator \cite{rojas2019variational}. There has been a recent study that presented the emerging robotics with considering safety and ergonomics \cite{gualtieri2021emerging}. Also, some studies emphasize more on the human factor in future robotics and automation with the transition to industry 4.0 \cite{kadir2021human,tseng2021sustainable}.

 The safety from the robot side is already covered extensively \cite{bixby1991proactive,alami2006safe,colgate2008safety,vasic2013safety,rojas2019variational} e.g., jerk minimization. Certain researches, considering safety from the human side, have included human motion kinematics to the automation problem. For instance, Pedrocchi presented a network architecture using tracking position sensors on a human to avoid collisions with robots \cite{pedrocchi2013safe}. The basic idea was further developed through predictive algorithms to interact better and have safe cooperation between humans and robots \cite{zanchettin2015safety,unhelkar2018human,kanazawa2019adaptive,chan2020collision}. Bajcsy also developed a robust human motion prediction for mobile robots to include the intended goals of the users \cite{bajcsy2020robust}. However, in-depth human intention detection is challenging due to complex and integrated human motion classifications \cite{goutsu2018classification}.
\begin{figure}[t!]
	\centering
%	\vspace{3mm} %5mm vertical space	
	\includegraphics[width=2.8 in]{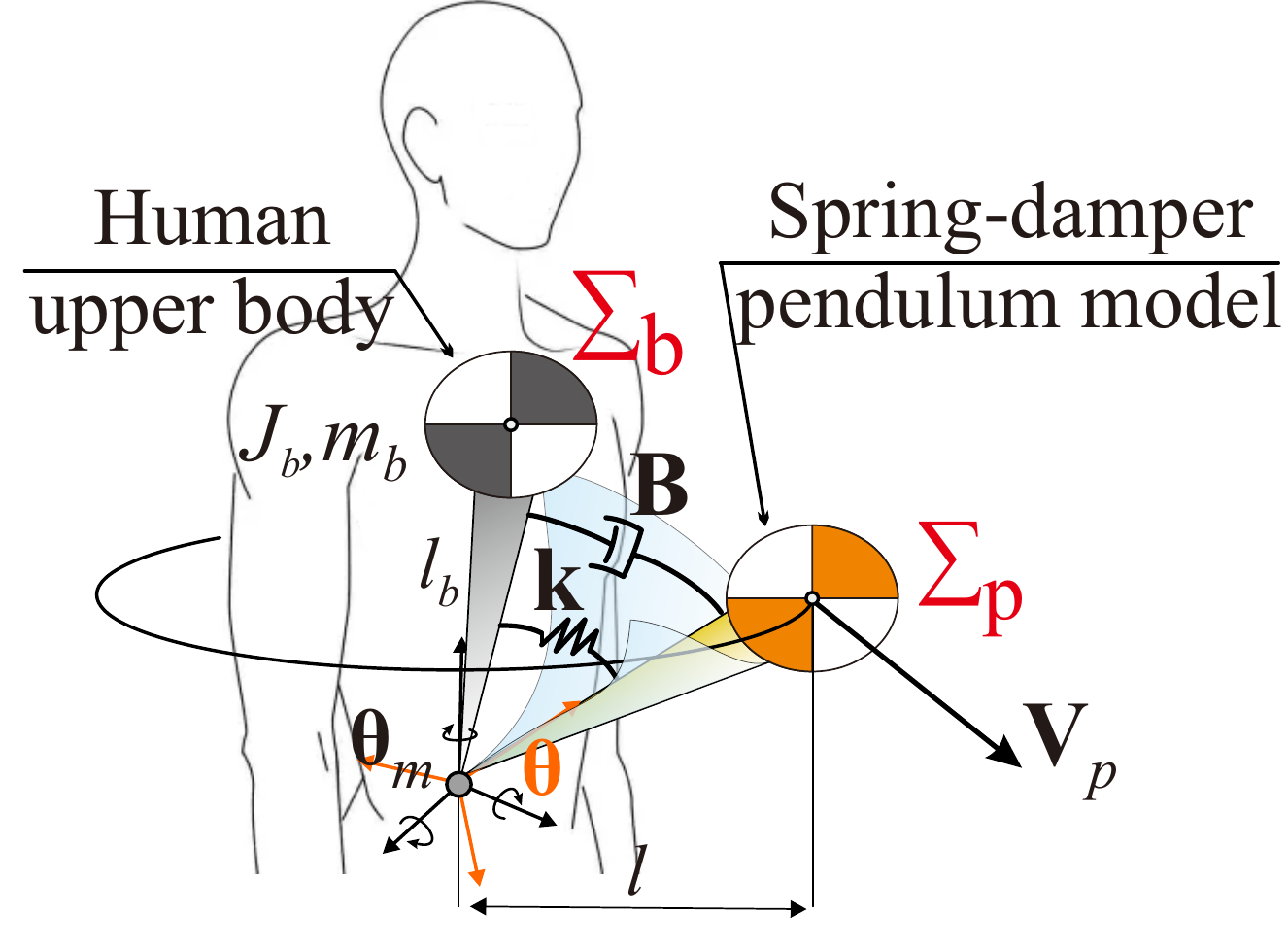}			
	\caption{A predictive safety model (PSM) with 3-Dof spring-damper pendulum system.}\label{Fig:Wheelchair_Model_3D}
\end{figure}

Considering the human counterpart as a black box creates certain limitations for assistive/cooperative robots and other guidance systems \cite{pedrocchi2013safe,zanchettin2015safety,unhelkar2018human,kanazawa2019adaptive,chan2020collision}. This open problem from the human point of view prevents the robots from getting into human's daily life. This issue mainly goes back to determining the human intention and real-time quantification of one's movements in the environment \cite{kratzer2020prediction,vianello2021human} that was shown in studies for human posture prediction for human limbs. However, these studies could have limitations due to applied leader-following strategies. Also, the large computations would make it impractical for real-time application \cite{shafti2019real,vianello2021human}. With our main motivation on the motion of the human upper body, we focus on how one moves safely with respect to certain physical references. Finding a way to quantify the safety of human orientation helps the robots to have information about the human quality of orientation and motion. In other words, quantifying the safety of human motion can bring systematic prediction methods to know possible anomalies in human motions. This can directly benefit the guidance and assistive robots to be more flexible and interactive workspace without fixed safeguards, in particular for disabled people or people who require continuous monitoring.

In this paper, we propose a new predictive safety model (PSM) that quantifies and recognizes the safe human orientation using a single inertial measurement unit (IMU) (see Fig. \ref{Fig:Wheelchair_Model_3D}). The open-source model utilizes both a dynamic reference dataset and connecting 3-Dof spring-damper pendulum model that predicts the safe motion of the human upper body\footnote{Open source directory: \href{https://github.com/SeyedAmirTafrishi/PredictiveSafetyModel}{github.com/SeyedAmirTafrishi/PredictiveSafetyModel} }. The proposed approach's contributions are as follows
\begin{itemize}
\item A dynamic and continuous quantification of the safety in the motion of the human upper body,

\item  Detection of safety in different levels from normal to unstable and unsafe orientation,

\item  Providing the physical proprieties as stiffness and damping values with application to different systems, e.g., assistive/guidance robots,

\item  Ability to evaluate the safety during complex human motions in real-time with a reduced-dimension safety dataset.
\end{itemize}

The paper is organized as follows. In Section \ref{Sec:DynamicModelSpringDamp}, the dynamic model of connecting 3-Dof spring-damper pendulum is derived. Next, the sensory information is explained in Section \ref{Sec:TheSensorDatacap}. In the rest of Section \ref{Sec:PredicitiveSafetyModel}, we explain the probability dataset design and variable estimation of the proposed safety pendulum model. We discuss the results in Section \ref{Sec:ResutlsandDISUCTION}. Finally, we conclude our findings in Section \ref{Sec:ConclusionD}.

\section{Dynamic Modeling of Spring-Damper Pendulum}
\label{Sec:DynamicModelSpringDamp}
To quantify the safety, we develop a new mathematical model utilizing a 3-Dof spring-damper pendulum. 
We have used the sketched model in Fig. \ref{Fig:Wheelchair_Model_3D} where the information of the human upper body motion on $\Sigma_b$ frame is available with the sensory system. Also, The 3-Dof spring-damper pendulum on $\Sigma_p$ frame refers to the pendulum model with the inertia that is connected to the human motion through a spring-damper system. The human motion on $\Sigma_b$ is compared with the predictive safety model $\Sigma_p$, and error presents how much human orientation and velocity deviate from normal and safe with respect to the standard safety dataset.
 
%% Kinematics 
At first, the position of the rotating pendulum are defined with respect to $\Sigma_p$ as follows
\begin{equation}
	\uvec{P}_p (t)=l_b\uvec{R}_y(\theta_y)\uvec{R}_x(\theta_x)=l_b\left[\begin{array}{c}
		\sin\theta_y(t) \cos\theta_x(t)\\
		\sin \theta_x(t)\\
		\cos\theta_y(t) \cos \theta_x(t)
	\end{array}\right]
	\label{Eq:thepositionofbody}
\end{equation}
where $\bm{\theta}=\{\theta_x,\theta_y,\theta_z\}$ are the angular orientation of the pendulum model and $l_b$ is the length from the human waist to the chest. Then, differentiating Eq. (\ref{Eq:thepositionofbody}) results the linear $\uvec{V}_p$ and angular $\bm{\omega}_p$ velocities as follows
\begin{align}
	&\uvec{V}_p=l_b \left( \dot{\theta}_x \left[\begin{array}{c}
		-\sin\theta_y \sin\theta_x\\
		\cos \theta_x\\
		-\cos\theta_y \sin \theta_x
	\end{array}\right] + \dot{\theta}_y \left[\begin{array}{c}
		\cos\theta_y \cos\theta_x\\
		0\\
		-\sin\theta_y \cos \theta_x
	\end{array}\right]\right) \nonumber\\
	&+l(t,\theta_x,\theta_y) \left[\begin{array}{c}
		0\\
		0\\
		\dot{\theta}_z
	\end{array}\right],\;\; \bm{\omega}_p=[\omega_x,\;\omega_y,\;\omega_z]^T
	\label{Eq:thevelocityofbody}
\end{align}
where $
	l(t,\theta_x,\theta_y) = \left[(\sin\theta_y \cos\theta_x)^2+(\sin\theta_x)^2\right]^{\frac{1}{2}}.$

%% Dynamics
We consider the system Lagrangian function by $L=E_T-E_U-E_P$ where $E_T$, $E_U$ and $E_P$ are the total kinetic and potential energies besides the Rayleigh dissipation function. Let us consider the potential energy is determined by the following two terms, gravity and virtual spring connected to human upper body, as follows
\begin{equation}{\small
E_U=m_bl_b g\uvec{P}^T_p \left[\begin{array}{c} 
		0\\
		0\\
		1
	\end{array} \right]+ U_k= m_b l_b g \cos\theta_y \cos \theta_x + U_k ,}
\end{equation}
where the energy of the virtual spring (stiffness) of the human body $U_k$ is 
\begin{eqnarray}
	U_k & = &	\frac{1}{2} \left(\bm{\theta}-\bm{\theta}_m \right)^T\uvec{K} \;\left(\bm{\theta}-\bm{\theta}_m \right) \nonumber\\
	&=&	\frac{1}{2}   \left[\begin{array}{ccc} 
		k_x & 0&0\\
		0& k_y &0\\
		0 &0  & k_z
	\end{array} \right] \left[\begin{array}{c} 
		(\theta_x-\theta_{m,x})^2\\
		(\theta_y-\theta_{m,y})^2\\
		({\theta}_z-{\theta}_{m,z})^2
	\end{array} \right] ,
\end{eqnarray}
where $\uvec{K}=\{k_{x},k_{y},k_z\}$ is the stiffness values in each axes and $\bm\theta_m$ is the estimated angles from the measurements of the sensors i.e., inertial measurement unit (IMU).
%where $d$ is the radius of body deviation from its central point that is found from
%\begin{equation}
%d=\frac{l_b}{2}[(\sin\theta_x\cos\theta_y)^2+(\sin\theta_x)^2]^{\frac{1}{2}}
%\end{equation}
Next, the total kinetic energy of the body is defined as follows
\begin{eqnarray}
	E_T&=&\frac{1}{2}m_b \uvec{V}^T_p\; \uvec{I}\; \uvec{V}_p  +\frac{1}{2} \bm{\omega}^T_p\uvec{J}_b \bm{\omega}_p   \nonumber\\
	&=& \frac{1}{2} m_b \Big [ l_b^2 \dot{\theta}^2_x+ l_b^2\dot{\theta}^2_y \cos^2\theta_x+l^2\dot{\theta}_z^2 \nonumber\\
	&-&2l_bl\dot{\theta}_x \dot{\theta}_z\cos\theta_y\sin\theta_x-2l_bl\dot{\theta}_y \dot{\theta}_z\sin\theta_y\cos\theta_x \Big ] \nonumber \\
	&+&\frac{1}{2}J_{b,x}\dot{\theta}^2_x  +\frac{1}{2}J_{b,y}\dot{\theta}^2_y +\frac{1}{2}J_{b,z}\dot{\theta}_z^2, 
\end{eqnarray}
Then, $E_P$ contains the virtual damper connected to the upper body that is defined by Rayleigh's dissipation function 
\begin{eqnarray}
 E_P&=& \frac{1}{2} \left(\dot{\bm{\theta}}-\dot{\bm{\theta}}_m \right)^T\uvec{B}\;\left(\dot{\bm{\theta}}-\dot{\bm{\theta}}_m\right) \nonumber\\
 & =&\frac{1}{2}   \left[\begin{array}{ccc} 
		b_x & 0&0\\
		0& b_y &0\\
		0 &0  & b_z
	\end{array} \right] \left[\begin{array}{c} 
		(\dot\theta_x-\dot\theta_{m,x})^2\\
		(\dot\theta_y-\dot\theta_{m,y})^2\\
		(\dot\theta_z-\dot\theta_{m,z})^2
	\end{array} \right]  
\end{eqnarray}
where $\uvec{B}=\{b_{x},b_{y},b_{z} \}$ is damping coefficient values in each axes. The Lagrangian function finally can be represented {\small 
\begin{align}%\frac{1}{2}J_{b,z}\dot{\theta}_z^2
	&L=\frac{1}{2} \Big ( m_b l_b^2 + J_{b,x}\Big )\dot{\theta}^2_x+\frac{1}{2}J_{b,z} \dot{\theta}_z^2 \nonumber\\
	&+\frac{1}{2}\Big ( m_b l_b^2 \cos^2\theta_x + J_{b,y}\Big )\dot{\theta}^2_y + m_b l_b g \cos\theta_y \cos \theta_x  \nonumber\\
	&+\frac{1}{2} m_b \Big [l^2\dot{\theta}_z^2-2l_bl\dot{\theta}_x \dot{\theta}_z \cos\theta_y\sin\theta_x-2l_bl\dot{\theta}_y \dot{\theta}_z \sin\theta_y\cos\theta_x \Big ]\nonumber\\
%	&-\frac{1}{2} \left(\bm{\theta}-\bm{\theta}_m \right)^T\uvec{K}\left(\bm{\theta}-\bm{\theta}_m \right)-\frac{1}{2} \left(\dot{\bm{\theta}}-\dot{\bm{\theta}}_m \right)^T\uvec{B}\left(\dot{\bm{\theta}}-\dot{\bm{\theta}}_m \right)
	& + \frac{1}{2} \left[b_{x} \left( \dot{\theta}_x- \dot{\theta}_{m,x}\right)^2+b_{y} \left( \dot{\theta}_y - \dot{\theta}_{m,y}\right)^2+ b_z \left(\dot{\theta}_z-\dot{\theta}_{m,z}\right)^2 \right]\nonumber\\
	&+\frac{1}{2} \left[ k_{x} \left(\theta_x-\theta_{m,x} \right)^2+k_{y} \left(\theta_y-\theta_{m,y} \right)^2+k_{z} \left(\theta_z-\theta_{m,z} \right)^2 \right].
	\label{Eq:Finallagrangianfunction}
\end{align}}

Next, the Lagrangian equations can be derived as follows for each $q$ configuration
\begin{align}
	\dfrac{d}{dt}\left(\dfrac{\partial E_T}{\partial \dot{q}}\right)- \dfrac{\partial E_T}{\partial q}+\dfrac{\partial E_P}{\partial \dot{q}}+\dfrac{\partial E_U}{\partial q}= \tau.
	\label{Eq:LagrangianEquationmain}
\end{align}
By applying (\ref{Eq:LagrangianEquationmain}) to the derived Lagrangian function $L$ in (\ref{Eq:Finallagrangianfunction}), we have following model
\begin{align}
 &	\uvec{M}(\bm{\theta}) \; \bm{\ddot{\theta}}+\uvec{H}(\bm{\theta},\bm{\dot{\theta}})+\uvec{G}(\bm\theta)=\bm{\tau}
%+\uvec{K}\left(\bm{\theta}-\bm{\theta}_m \right)+\uvec{K}'_{\theta}\left(\bm{\theta}-\bm{\theta}_m \right)\nonumber\\
%	&\uvec{B}\left(\dot{\bm{\theta}}-\dot{\bm{\theta}}_m \right)+\uvec{B}'_{\dot{\theta}}\left(\dot{\bm{\theta}}-\dot{\bm{\theta}}_m \right)=\uvec{u},
	\label{Eq:MotionEquationGeneral}
\end{align}
where states are $\bm{\theta}=\{\theta_x,\theta_y,\theta_z\}$ and terms are  
\begin{align}
	&	\uvec{M}= \left[\begin{array}{ccc} {M}_{11} &  {M}_{12}& {M}_{13}\\ {M}_{21} &  {M}_{22}&  {M}_{23}\\ {M}_{31}  &  {M}_{32}  &  {M}_{33}  \end{array}\right], \uvec{H}= \left[\begin{array}{cccc} {h}_1& {h}_2  & {h}_3 \end{array}\right]^T, \nonumber \\
	&\uvec{G}=\left[\begin{array}{cccc} {g}_{1}& {g}_{2}&  {g}_{3}  \end{array}\right]^T,\;\bm{\tau}=\left[\begin{array}{cccc} {\tau}_{x}& {\tau}_{y}&  {\tau}_{z}  \end{array}\right]^T,
	\label{Eq:MatricesMotionforward}
\end{align}
where
\begin{align}
	&M_{11}= m_b l^2_b + J_{b,x} , \; M_{12}= M_{21}= 0,\;  \nonumber\\
	& M_{13} = M_{31} = -l_b l \cos\theta_y \sin \theta_x, M_{33}= m_b l^2 + J_{b,z} \nonumber \\
	& M_{22}=  m_b l^2_b  \cos^2\theta_x + J_{b,y} ,\; M_{23}= M_{32}=-l_b l \sin \theta_y \cos \theta_x , 
\end{align}
and
\begin{align}
	&h_1= m_b l^2_b \dot{\theta}^2_y \sin \theta_x \cos \theta_y + k_{x} \left( \theta_x-\theta_{m,x} \right)  \nonumber \\	
	&+ b_{x}\left( \dot{\theta}_x- \dot{\theta}_{m,x} \right)+m_b\Big[-l_b \dot{l} \dot{\theta}_z  \cos\theta_y\sin\theta_x -\Big( l l'_{x} \dot{\theta}_z^2 \nonumber\\  
	&- l_b l'_{x}\dot{\theta}_x \dot{\theta}_z \cos \theta_y \sin \theta_x - l_b l'_x \dot{\theta}_y \dot{\theta}_z \sin \theta_y \cos \theta_x    \Big) \Big] \nonumber \\
   &	h_2 =   - m_b l^2_b \dot\theta_x \dot\theta_y \sin 2 \theta_x  + k_{y} \left( \theta_y-\theta_{m,y} \right)  \nonumber \\	
	&+  b_{y}\left( \dot{\theta}_y- \dot{\theta}_{m,y} \right)+m_b\Big[-l_b \dot{l} \dot{\theta}_z   \sin\theta_y\cos\theta_x -\Big( l l'_{y} \dot{\theta}_z^2 \nonumber\\  
	&-  l_b l'_{y}\dot{\theta}_x \dot{\theta}_z \cos \theta_y \sin \theta_x - l_b l'_y \dot{\theta}_y \dot{\theta}_z \sin \theta_y \cos \theta_x    \Big) \Big] \nonumber  \\
	&h_3=  - m_b l_b l  \left(\dot{\theta}^2_x+\dot\theta^2_y\right) \cos \theta_x \cos \theta_y + k_{z} \left( \theta_z-\theta_{m,z} \right)  \nonumber \\	
	&+  b_{z}\left( \dot{\theta}_z- \dot{\theta}_{m,z} \right)+ 2 m_b  l_b l \dot\theta_x \dot\theta_y \sin \theta_x \sin \theta_y  \nonumber\\
	& -   m_b \Big [ l_b \dot{l} \dot{\theta}_x \cos \theta_y \sin \theta_x + l_b \dot{l} \dot{\theta}_y \sin \theta_y \cos \theta_x    \Big], \nonumber
\end{align}
\begin{align}
	& g_1=  m_b l_b g \sin \theta_x \cos \theta_y ,\; g_2 =  m_b l_b g \cos \theta_x \sin \theta_y ,\;g_3=0 \nonumber\\	
\end{align}
\begin{figure}[t!]
	\centering
	\vspace{3mm} %5mm vertical space	
	\includegraphics[width=3.4 in]{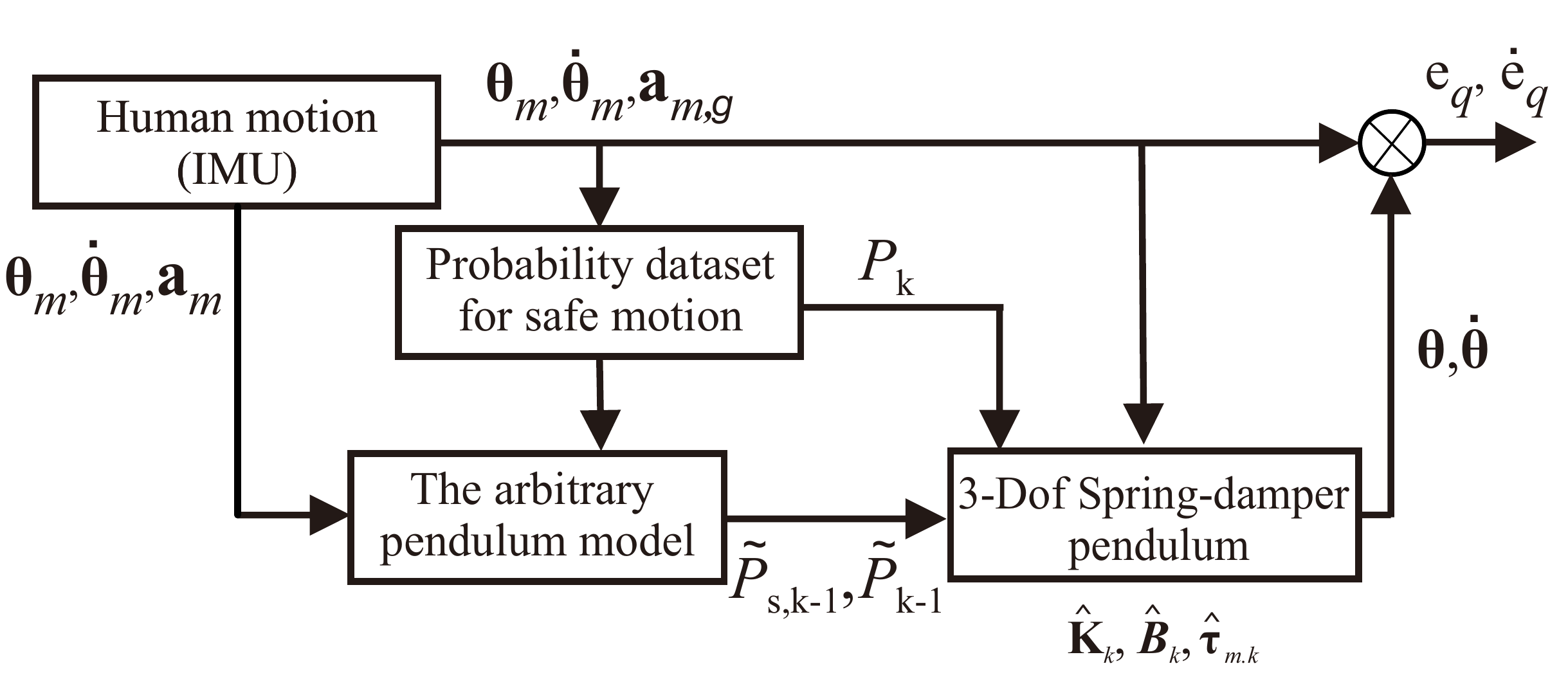}			
	\caption{The block diagram of the predictive safety model (PSM).}\label{Fig:SafeBlockDiagrams}
\end{figure}
while derivatives of the changing distance of the pendulum bob $l$ is derived by (\ref{Eq:thepositionofbody}) as  
{\small
	\begin{align}
		& l(t)= \left( \sin^2 \theta_y \cos^2 \theta_x + \sin^2\theta_x \right)^{\frac{1}{2}}, \nonumber\\
		& \dot{l}(t)=\frac{1}{2}\left( \dot{\theta}_y \sin 2 \theta_y \cos^2 \theta_x + \dot{\theta}_x \sin 2 \theta_x \sin^2 \theta_y \right) \nonumber \\
		& \cdot \left(\sin^2 \theta_y \cos^2 \theta_x + \sin^2\theta_x \right)^{-\frac{1}{2}}, \nonumber \\
		&  l'_x(t)=\frac{\partial l}{\partial \theta_x} = \frac{1}{2}\left(\sin 2 \theta_x \cos^2 \theta_y \right)\left(\sin^2\theta_y \cos^2 \theta_x + \sin^2\theta_x\right)^{-\frac{1}{2}}, \nonumber \\ 
		&  l'_y(t)=\frac{\partial l}{\partial \theta_y} = \frac{1}{2}\left(\sin 2 \theta_y \cos^2 \theta_x \right)\left(\sin^2\theta_y \cos^2 \theta_x + \sin^2\theta_x\right)^{-\frac{1}{2}}.
\end{align}}

It is important to note that connecting spring-damper coefficients $\{\uvec{K},\;\uvec{B}\}$ are assumed to be independent from variables ($\bm\theta,\dot{\bm{\theta}}$) of the pendulum model. Also, the variables $\{\uvec{K}, \uvec{B}, \bm{\tau}\}$ in motion equation (\ref{Eq:MotionEquationGeneral}) will be the main factors in estimating the safe orientation of the human in incoming sections.
\begin{figure}[t!]
	\centering
	\vspace{3mm} %5mm vertical space	
	\includegraphics[width=2.4 in]{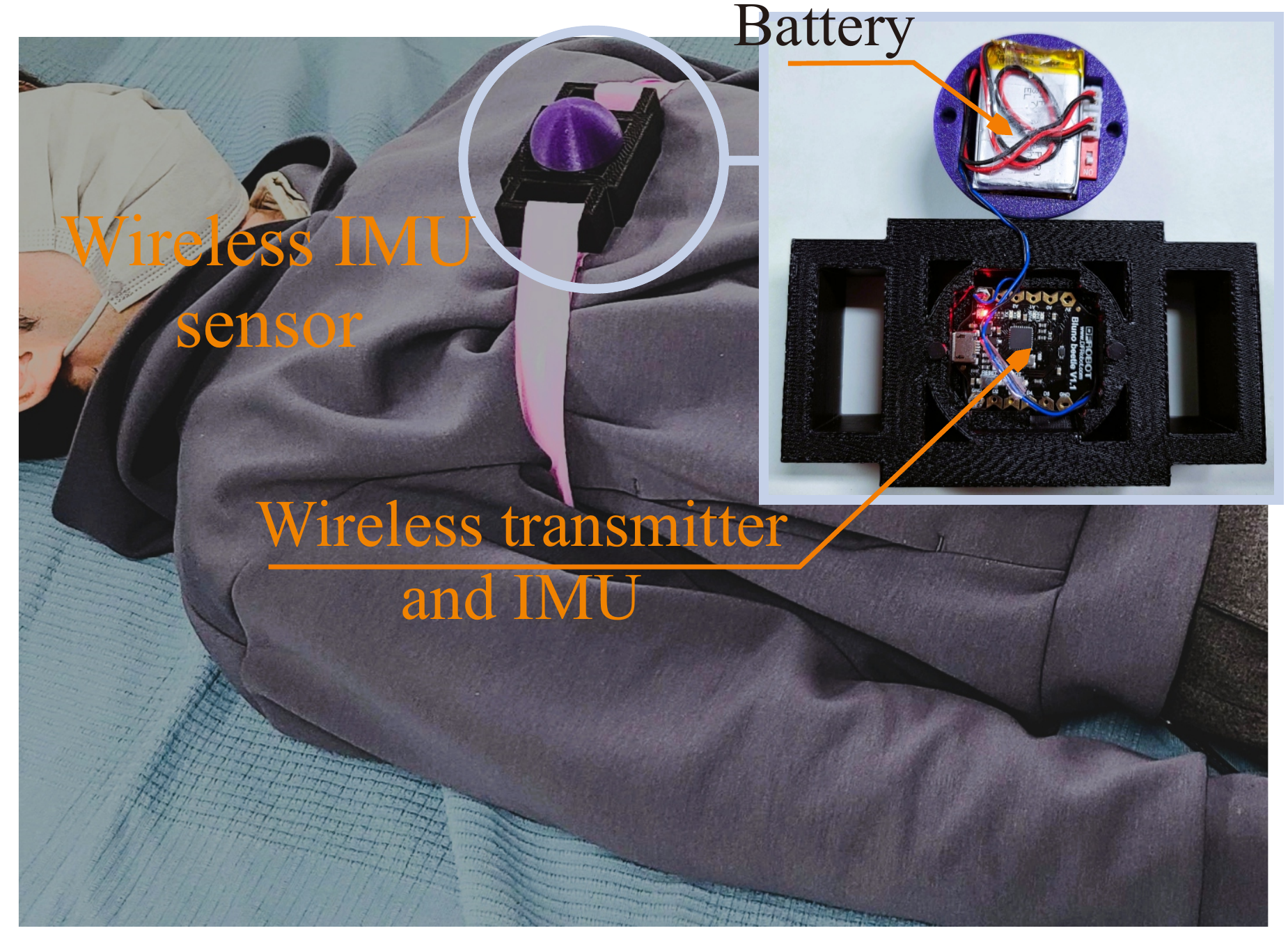}				
	\caption{The designed wireless IMU sensor is attached to the chest of participants.}\label{Fig:IMU_Chest}
\end{figure}

\section{Predictive Safety Model}
\label{Sec:PredicitiveSafetyModel}
 The aim is to develop a model that can give a predictive safe motion of human's upper body orientation. Fig. \ref{Fig:SafeBlockDiagrams} shows the block diagram of how the safety model is developed.  The measured data from the inertial measurement unit (IMU) is first used to calculate probability parameters from a dataset. Then, the physical parameters are used in the arbitrary pendulum model. Finally, the 3-Dof the spring-damper pendulum uses the estimated variables that are the function of processed data as well as the direct inputs from IMU sensor. 

\subsection{Captured Sensor Data}
\label{Sec:TheSensorDatacap}
The inertial measurement unit is the only sensor utilized to understand human orientation, angular velocity and approximated linear acceleration. As shown in Fig. \ref{Fig:IMU_Chest}, a designed wireless IMU sensor is fixed on the chest of the participant, located on $\Sigma_b$ frame.

We have developed a wireless module with IMU sensor, i.e., Bno055, that encompass accelerometer, gyroscope and magnetometer. An AHRS filter is applied \cite{roetenberg2005compensation,Tafrishi2021Sen} to obtain the estimated orientation and angular velocity. In each iteration, the filter estimates orientation $\bm{\theta}_m=\left[\theta_{m,x},\; \theta_{m,y},\; \theta_{m,z}\right]^T$ and angular velocity without bias ${\footnotesize\dot{\bm{\theta}}_m=\left[\dot{\theta}_{m,x}, \;\dot{\theta}_{m,y},\;\dot{\theta}_{m,z}\right]^T}$. Also, the acceleration of human upper body is another variable that will be measured from accelerometer. Because of existing high noise, the measured raw acceleration is transformed into s-domain $\uvec{A}_c$ and is filtered by zero-phase low-pass filter %(signal processing toolbox of Matlab) 
$\uvec{f}$ as follows \cite{Tafrishi2021Sen,mitra2006digital}
\begin{equation}
	\uvec{A}_m(s)= \uvec{f}(n_f,f_c) \; \uvec{A}_c(s),
	\label{Eq:filterbutterworth}
\end{equation}
where $n_f$ and $f_c$ are the order and cut-off frequency ratio of low-pass Butterworth filter. Then, we can obtain the filtered signal $\uvec{a}_m(t)=[a_{x,m}, \; a_{y,m}, \; a_{z,m}]^T$ in time domain for acceleration from (\ref{Eq:filterbutterworth}). In our model, it is required to know the angular direction of the exerted force $\uvec{F}_m= m_b\;\uvec{a}_m$ on moving human body (captured by accelerometer); hence, the acceleration excluding gravitation force is computed as follows 
\begin{equation}
	\uvec{a}_{m,g}= \uvec{R}(t) \; \uvec{a}_m (t) - \uvec{R} (t_0) \; \uvec{a}_g,
\end{equation}
where $\uvec{a}_g$ is the initial measurement in time $t_0$ that presents the gravitational force direction and $\uvec{R}(t)=\uvec{R}(\theta_z)\uvec{R}(\theta_y)\uvec{R}(\theta_x)$ is the yaw-pitch-roll rotation matrix with respect to global frame. After this computation, the angular direction of exerted force on body is derived by 
\begin{align}
 &	\bm{\varphi}=  \tan^{-1} \frac{\uvec{a}_{m,g}}{\parallel \uvec{a}_{m,g} \parallel}= \Bigg [\tan^{-1}\left(\frac{a_{x,m}}{\parallel \uvec{a}_{m,g} \parallel}\right),\nonumber\\
 & \tan^{-1}\left(\frac{a_{y,m}}{\parallel \uvec{a}_{m,g} \parallel}\right), \tan^{-1}\left(\frac{a_{z,m}}{\parallel \uvec{a}_{m,g}\parallel}\right)\Bigg]^T.   
 \label{Eq:Theangularextertedforce}
\end{align}

\subsection{Reduced-Dimension Probability Dataset of Safety}
\label{Sec:DatasetProblity}
\begin{figure}[t!]
	\centering
	\vspace{3mm} %5mm vertical space	
	\includegraphics[width=2.8 in]{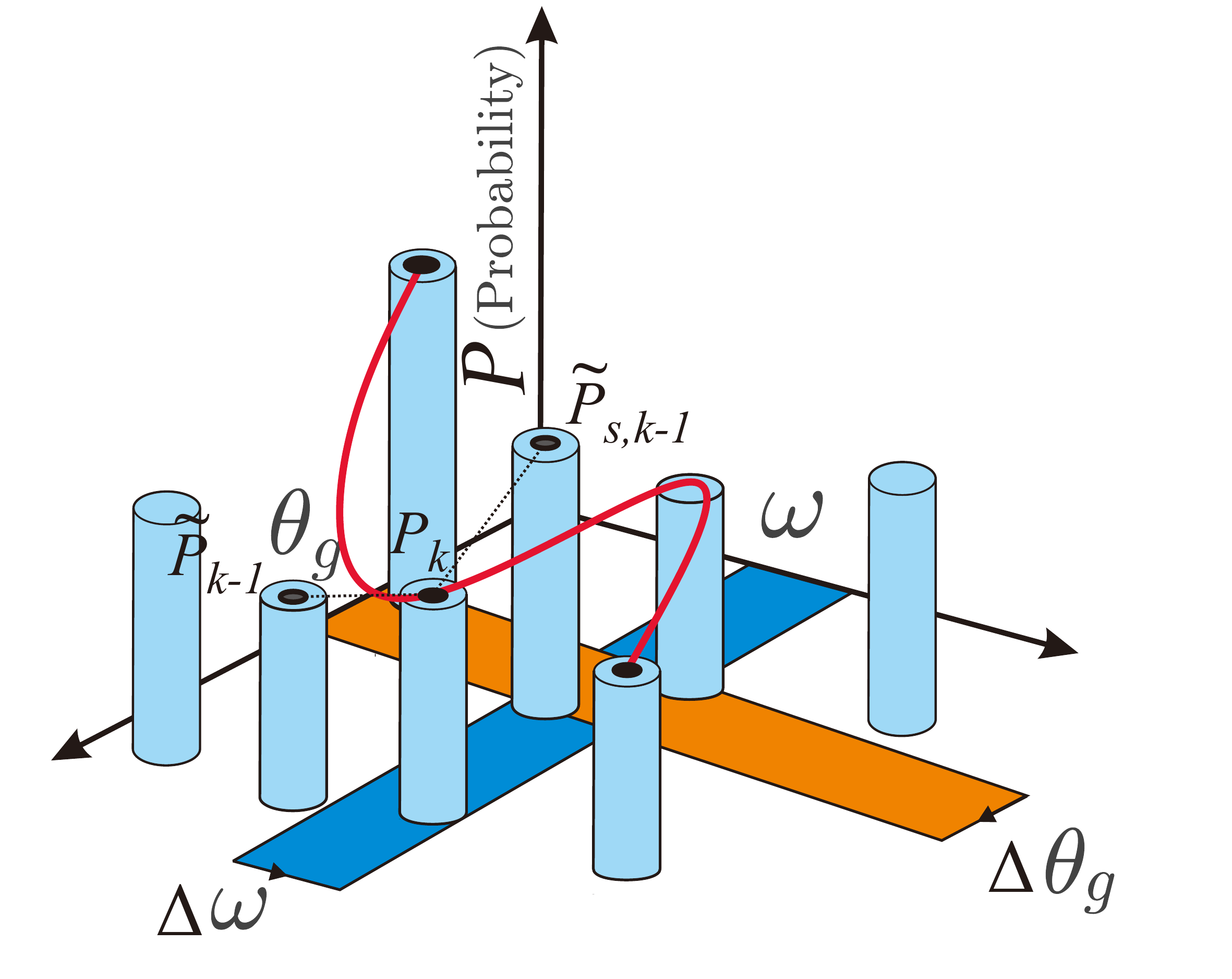}		
	\caption{Prediction by using the reduced-dimension probability datasets. Note that $P_k$ is the current step's probability, and $\tilde{P}_{k-1}$ and $\tilde{P}_{s,k-1}$ are the predicted probability and the probability of predicted safe motion  from step $k-1$.}\label{Fig:TheProbCap1}
\end{figure}

The dataset to determine the safest motion of the human upper-body orientation is essential. From previous studies \cite{gavrila1999visual,najafi2003ambulatory,prakash2018recent,goutsu2018classification}, we already know determining human motion intention or gait in different body configurations is challenging. Human tends to make many complex motions. Although not all the motions can be exactly determined, a strategy to summarize the captured data from different participants can be a solution in finding whether the human/patient is following the correct motion.

In this study, we propose a new dimension reduction technique from the recorded data by considering three main factors, as shown in Fig. \ref{Fig:TheProbCap1}. The angular direction of the body $\theta_g$ with respect to the gravity, the norm of body's angular velocity $\omega= \parallel \dot{\bm{\theta}}_m \parallel \in \left[0, \; \omega_{max} \right]$ and probability of each action based on recorded $\left( \theta_g, \omega \right)$ are the main contributing parameters in reduced-dimension safety dataset.  The angular direction of the body $\theta_g$ with respect to the gravity and the norm of the body's angular velocity $\omega$ give enough information to formalize our prediction model. In a simple physical sense, the reason is that we can see the overall force on the mass of inertia (human upper-body) with respect to gravity and its momentum in time. This method reduces a six-dimensional complex problem to a one-dimensional system. Later, these parameters will be projected back and used to estimate the variables $\{\uvec{K},\;\uvec{B},\;\bm{\tau}\}$ in the designed spring-damper pendulum as the safety model.

\begin{figure}[t!]
	\centering
	\vspace{3mm} %5mm vertical space	
	\includegraphics[width=1.8 in]{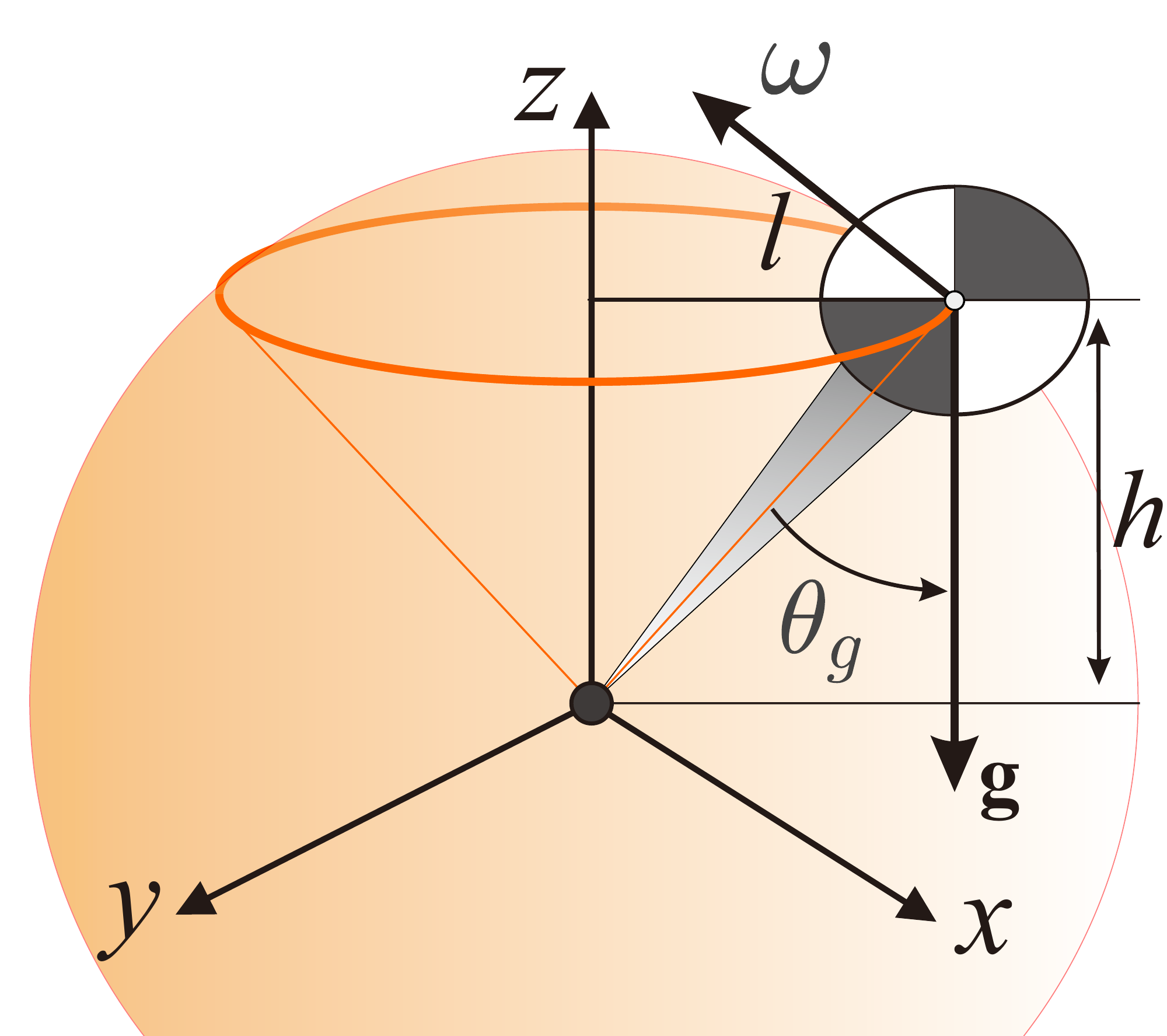}				
	\caption{The variables in the reduced-dimension (1-Dof) arbitrary pendulum model.}\label{Fig:TheMotionGravity}
\end{figure}
To compute the angular direction of the body $\theta_g \in \left[\theta_{g,min}, \theta_{g,max} \right]$ with respect to gravity (see Fig. \ref{Fig:TheMotionGravity}), we can use trigonometric relations from (\ref{Eq:thepositionofbody}) as follows  
\begin{equation*}
	\theta_g=\begin{cases}
		&\pi-2\cos^{-1}\left(h/a_c\right),\;\;\;\;\;\;\;\;\;\;\;\;\;\;\;\;\;\;\;\;h \leq  l_b\\
		& (\pi/2)-\sin^{-1}\left((h-l_b)/l_b\right),\;\;\;\;\;\;\;\;\hfill h >  l_b
	\end{cases}
\end{equation*}
where 
$
 h= l_b \left[1- \cos \theta_x \cos \theta_y \right],\; a_c=(h^2+l^2)^{\frac{1}{2}}.
$

Furthermore, the probability distribution in the matrix $\uvec{P}_{n,m} \in \mathbb{R}^2 \rightarrow \mathbb{R}^{n\times m} $ is defined as follows
\begin{equation}
	\uvec{P}_{n,m} = \left[\begin{array}{cccc}
		P_{1,1} &P_{1,2}	& \cdots  & P_{1,m} \\
		\vdots &    &	\ddots & \\
		P_{n,1}	 & P_{n,2} &  \cdots   &  P_{n,m}
	\end{array}\right]
\label{Eq:Theprobilitysafematrixara}
\end{equation}
where each element corresponds to the probability of the motion based on $(\theta_g,\omega)$ variables. Also, the matrix $\uvec{P}_{n,m}$ element in (\ref{Eq:Theprobilitysafematrixara}) corresponds to following information of $(\theta_{g,k},\omega_k)$ in $k$-th sample by ($n_k,m_k$) as 
\begin{equation}
\theta_{g,k}=\left(\theta_{g,\max}-\theta_{g,\min}\right)\frac{n_k}{n_{\theta}}+\theta_{g,\min},\;	\omega_k=\omega_{max} \frac{m_k}{m_\omega},
\end{equation}
where $m_{\omega}$ and $n_{\theta}$ are the maximum number of arrays for the norm of angular velocity $\omega$ and angular direction of gravity force $\theta_g$. It is important note that the distribution continuity for matching $(\theta_g,\omega)$ in $\uvec{P}_{n,m}$ depends on the accuracy range of angular direction of gravity $\varepsilon_{\theta}=1/n_{\theta}$ and accuracy range of the norm of angular velocity $\varepsilon_\omega=1/m_{\omega}$. The accuracy range is an important part of the determination for probability distribution matrix $\uvec{P}_{n,m}$ since the probability of each value may include different motions. This means high accuracy might create more smooth motions but it will make the proper evaluation of the distinguished motion patterns harder. In the incoming section, we find a proper way to develop a prediction model and estimate the variables of the safety pendulum model through the collected dataset.

\alglanguage{pseudocode}
\begin{algorithm}[t!] 
\caption{The computation of safety model.}
\begin{algorithmic}[1]
\Procedure{PredictProc}{$\bm{\theta}_m,\dot{\bm{\theta}}_m,\uvec{P}_{n,m}$}
%\While {$e_s \geq \epsilon_s $} \Comment{Phase III }
\State Compute $\theta_{g,k}$, $\omega_k$ and $P_k$
\State Determine the predicted $\tilde{\theta}_{g,k-1}$ and $\tilde{\omega}_{k-1}$ by (\ref{Eq:Thepredictedstepvariables})
\State Calculate the probability of predicted motion $\tilde{P}_{k-1}$ from $k-1$ step according to (\ref{Eq:safestproblityargument}) 
\State Find $\uvec{P}_{n',m'}$ from (\ref{Eq:Theprobilitysafematrixhorizen}) 
\If {$ \left| \uvec{P}_{n',m'} \right| \leq \varepsilon_p $} 
\State Update $\tilde{P}_{s,k-1}$ from $\tilde{P}_{s,k-2}$ by ($\Delta \theta_{g,k}, \Delta \omega_k$) step
%$\tilde{P}_{s,k-1} \leftarrow \tilde{P}_{s,k-2}$ % according to (\ref{Eq:Rupdateofphase2})
\Else
\State Find highest probability for safe motion $\tilde{P}_{s,k-1}$ from $\uvec{P}_{n',m'}$ from (\ref{Eq:safestproblityargument})
\EndIf
\State Estimate spring-damper pendulum variables $\hat{\uvec{K}}_k$, $\hat{\uvec{B}}_k$ and $\hat{\bm{\tau}}_{m,k}$ from (\ref{Eq:TheDampingmatrixupdates}) and (\ref{Eq:Thetorqueestimateupdate})
\State Solve the spring-damper pendulum model in (\ref{Eq:MotionEquationGeneral}) 
\State Calculate the orientation and velocity safety deviation from (\ref{Eq:Safetyformulationcase})

%\EndWhile
\Return $e_{\theta}(k),e_{\omega}(k), E_{m,\theta}(k)$ and $E_{m,\omega}(k)$
\EndProcedure
\end{algorithmic}
\label{Algo:processsafetymodel}
\end{algorithm}\alglanguage{pseudocode}

\subsection{Variable Estimation of Safety Pendulum Model}
 After having a proper safety dataset for predicting the parameters, variables of the spring-damper pendulum model will be estimated. The algorithm \ref{Algo:processsafetymodel} presents the process of the computation for the safety pendulum model that is demonstrated in Fig \ref{Fig:SafeBlockDiagrams}.

%((Jb+Mbw*lb^2)*((OmegaK-OmegaSafekn)/(Tnew-told(end)))+Mbw*lb*sin(ThetaGsafek));

The safe motion with connected spring-damper model (\ref{Eq:MotionEquationGeneral}) are defined with the estimated stiffness and damping coefficients at $k$-th sample as follows
\begin{equation}	
	 \hat{\uvec{K}}_k=\uvec{K}_c+\hat{k}_{s,k} \uvec{I},\;\;
     \hat{\uvec{B}}_k=\uvec{B}_c+ \hat{b}_{s,k} \uvec{I}.
     \label{Eq:TheDampingmatrixupdates}
\end{equation}
where $\uvec{K}_c$ and $\uvec{B}_c$ are human body stiffness and damping constants and estimated safe stiffness and damping coef. are
\begin{equation}
	 \hat{k}_{s,k}\triangleq \frac{\hat{F}_k}{\hat{\theta}_{g,k}}\frac{1}{(1-P_k)^2},\;\;\;
	 \hat{b}_{s,k}\triangleq \frac{\hat{F}_k}{\hat{\omega}_{k}}\frac{2}{(1-P_k)^3}.
	 \label{Eq:Thedampingcoefficientsesti}
\end{equation}
The estimated variables in (\ref{Eq:Thedampingcoefficientsesti}) are defined with linear estimate approximation and reduced-dimension pendulum model (Fig. \ref{Fig:TheMotionGravity}) in the following form
\begin{align} 
	\label{Eq:Estimatedforcek}
	 & \hat{F}_k = \frac{1}{l_b}\Big[\frac{1}{T}(J_{b,s}+m_b l_b^2)\left(\hat{\omega}_k-\tilde{\omega}_{s,k-1}\right)  + & m_b g l_b \sin \hat{\theta}_{gs,k} \Big], \\
	& \hat{\theta}_{g,k} =   \tilde{\theta}_{g,k-1}-\left(\theta_{g,k}-\tilde{\theta}_{s,k-1}\right),\\ 
    &  \hat{\omega}_{k} = \tilde{\omega}_{k-1}-\left(\omega_k - \tilde{\omega}_{s,k-1} \right),
      \label{Eq:Estimateedangularvelk}
\end{align}
where $T$, $J_{b,s}\in \mathbb{R}$ and $\{\tilde{\theta}_{s,k-1},\tilde{\omega}_{s,k-1}\}$ are the sampling time and approximated inertia from $\uvec{J}_b$ and the predicted safe angular orientation and velocities. 
The estimated torque for the spring-damper pendulum $\hat{\bm{\tau}}_{m,k}$ is derived by using the angular direction of exerted acceleration $\bm{\varphi}$ from the IMU sensor  (\ref{Eq:Theangularextertedforce}) that is 
\begin{equation}
		\hat{\bm{\tau}}_{m,k}=(l_b\uvec{F}_m+\uvec{G}(\bm{\theta_m}))+\left[(l_b \uvec{F}_m+\uvec{G}(\bm{\theta_m}))-l_b \hat{F}_k \tan \bm{\varphi}\right].
		\label{Eq:Thetorqueestimateupdate}
\end{equation}
Note that Eq. (\ref{Eq:Thetorqueestimateupdate}) has gravity term $\uvec{G}(\bm{\theta}_m)$ with substituted sensory information of angular orientation to give correct torque to the safety model (\ref{Eq:MotionEquationGeneral}). Also, the estimated force $\hat{F}_k$ in $k$-th sample is designed based on a pendulum with 1-Dof arbitrary orientation as shown in Fig. \ref{Fig:TheProbCap1}. This pendulum model is considered with inertia and gravitational force that result in (\ref{Eq:Estimatedforcek}) from  
\begin{equation}
\hat{\tau}_k=l_b \hat{F}_k=(J_{b,s}+m_b l_b^2) \hat{\ddot{q}}_k+m_bgh_k
\end{equation}
where 
\begin{equation}
\hat{\ddot{q}}_k =\frac{1}{T}\left(\hat{\omega}_k-\tilde{\omega}_{s,k-1}\right) ,\; h_k= l_b \sin \hat{\theta}_{gs,k}.
\end{equation}

To determine the predicted safety variables $(\tilde{\theta}_{g,k-1},\tilde{\omega}_{k-1})$ in (\ref{Eq:Estimatedforcek})-(\ref{Eq:Estimateedangularvelk}), we find them in following updates
\begin{eqnarray}
 \left[\begin{array}{c}
	\tilde{\theta}_{g,k-1} \\
    \tilde{\omega}_{k-1}
\end{array} \right] =  \left[\begin{array}{c}
\theta_{g,k-1}+\Delta\theta_{g,k} \\
\omega_{k-1}+\Delta\omega_{k}
\end{array} \right],
\label{Eq:Thepredictedstepvariables}
\end{eqnarray}
where step updates are computed by
\begin{equation*}
	\Delta\theta_{g,k} = \theta_{g,k}-\theta_{g,k-1},\; \Delta\omega_{k}=\omega_{k}-\omega_{k-1}.
\end{equation*}

Then, we can derive the predicted probability $\tilde{P}_{k-1}$ and its values for $\{\theta_{s,k-1},\omega_{s,k-1}\}$ from previous step $k-1$ by using the change of the angular orientation and norm of angular velocity as follows 
\begin{align}
	\tilde{P}_{k-1} = {P}_{\tilde{n},\tilde{m}} (\theta_{g,k},\omega_k)
\end{align}
where 
\begin{align}
&\tilde{n}= \begin{cases}
	&\left\lceil \frac{ (\theta_{g,k-1}+\Delta\theta_{g,k})n_{\theta}}{(\theta_{g,{max}}-\theta_{g,{min}} )} \right\rceil,  \;\;\;\;\;\;\;\;\; \Delta\theta_{g,k}>0 \\
	& {\footnotesize} \\
	&\left\lfloor   \frac{ (\theta_{g,k-1}+\Delta\theta_{g,k})n_{\theta}}{(\theta_{g,{max}}-\theta_{g,{min}} )} \right\rfloor, \;\;\;\;\;\;\;\;\; \Delta\theta_{g,k} \leq 0
\end{cases} ,\\ 
&\tilde{m}= \begin{cases} & \left\lceil \frac{(\omega_{k-1}+\Delta\omega_k)m_\omega}{\omega_{max}} \right\rceil , \;\;\;\;\;\;\;\;\;\;\; \Delta\omega_k>0 \\
	& {\footnotesize}\\
	& \left\lfloor \frac{ (\omega_{k-1}+\Delta\omega_k)m_\omega}{\omega_{max}} \right\rfloor, \;\;\;\;\;\;\;\;\;\;\; \Delta\omega_k \leq 0 
	\end{cases}
\end{align} % \Delta\theta_g
\begin{figure}[t!]
	\centering
	\vspace{3mm} %5mm vertical space	
	\includegraphics[width=3.2 in]{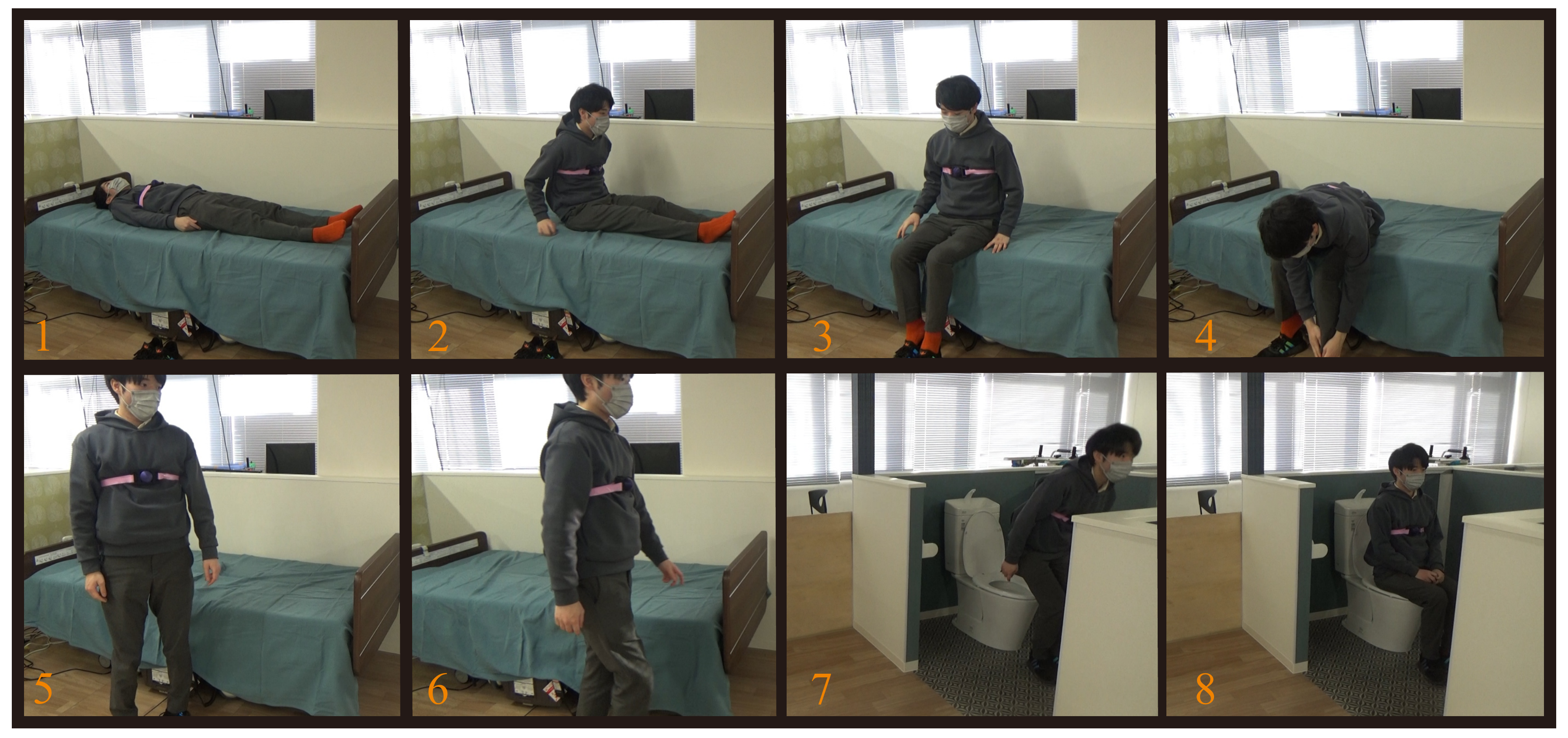}	
	\caption{The chosen scenario for the experiment at the Living lab of Tohoku University.}\label{Fig:Scenario_Paper_Safety}
\end{figure}
Also, the predicted probability $\tilde{P}_{s,k-1}$ for the safe motion in $k$-th sample is defined by 
\begin{equation}
	\tilde{P}_{s,k-1} =\underset{P_k}{\arg \max}  \{ \tilde{\uvec{P}}_{n',m'} \},
	\label{Eq:safestproblityargument}
\end{equation}
where 
\begin{equation}
	\uvec{P}_{n',m'}= \left[\begin{array}{cc}
		P_{n'_f m'_f} & P_{n'_f,m'_c}\\
		P_{n'_c,m'_f} & P_{n'_c,m'_c}
		\label{Eq:Theprobilitysafematrixhorizen}
	\end{array}\right]
\end{equation}
where row number with floor and ceil functions ${\footnotesize\left(n'_f,n'_c \right)}$ and column number with floor and ceil functions  ${\footnotesize\left(m'_f,m'_c \right)}$ for safe matrix candidate $\tilde{\uvec{P}}_{n',m'}$ of probability are 
\begin{align*}
&	n'_f = \Big\lfloor \frac{ \big(\tilde{\theta}_{g,k-1}+\theta_{g,{min}} \big)n_{\theta}}{\left( \theta_{g,{max}}-\theta_{g,{min}} \right)} \Big\rfloor ,    m'_f = \big \lfloor \tilde{\omega}_{k-1} \left(m_\omega/\omega_{max} \right)  \big\rfloor,\\	
&	n'_c = \Big\lceil \frac{ \big(\tilde{\theta}_{g,k-1}+\theta_{g,{min}} \big)n_{\theta}}{\left( \theta_{g,{max}}-\theta_{g,{min}} \right)}  \Big\rceil ,   m'_c = \big\lceil \tilde{\omega}_{k-1} \left(m_\omega/\omega_{max} \right)\big\rceil.
\end{align*}
This distribution of probability in $\uvec{P}'_{n,m}$ finds the safest path with the highest probability with an expanded horizon. It is important to note that here our horizon is 2 constant arrays with respect to $(\theta_g,\omega)$ variables. However, different algorithms with a larger horizon range can be utilized to find the safest and smoothest motions which is a separate study.

\begin{figure}[t!]
	\centering
	%\vspace{3mm} %5mm vertical space	
	\includegraphics[width=3 in]{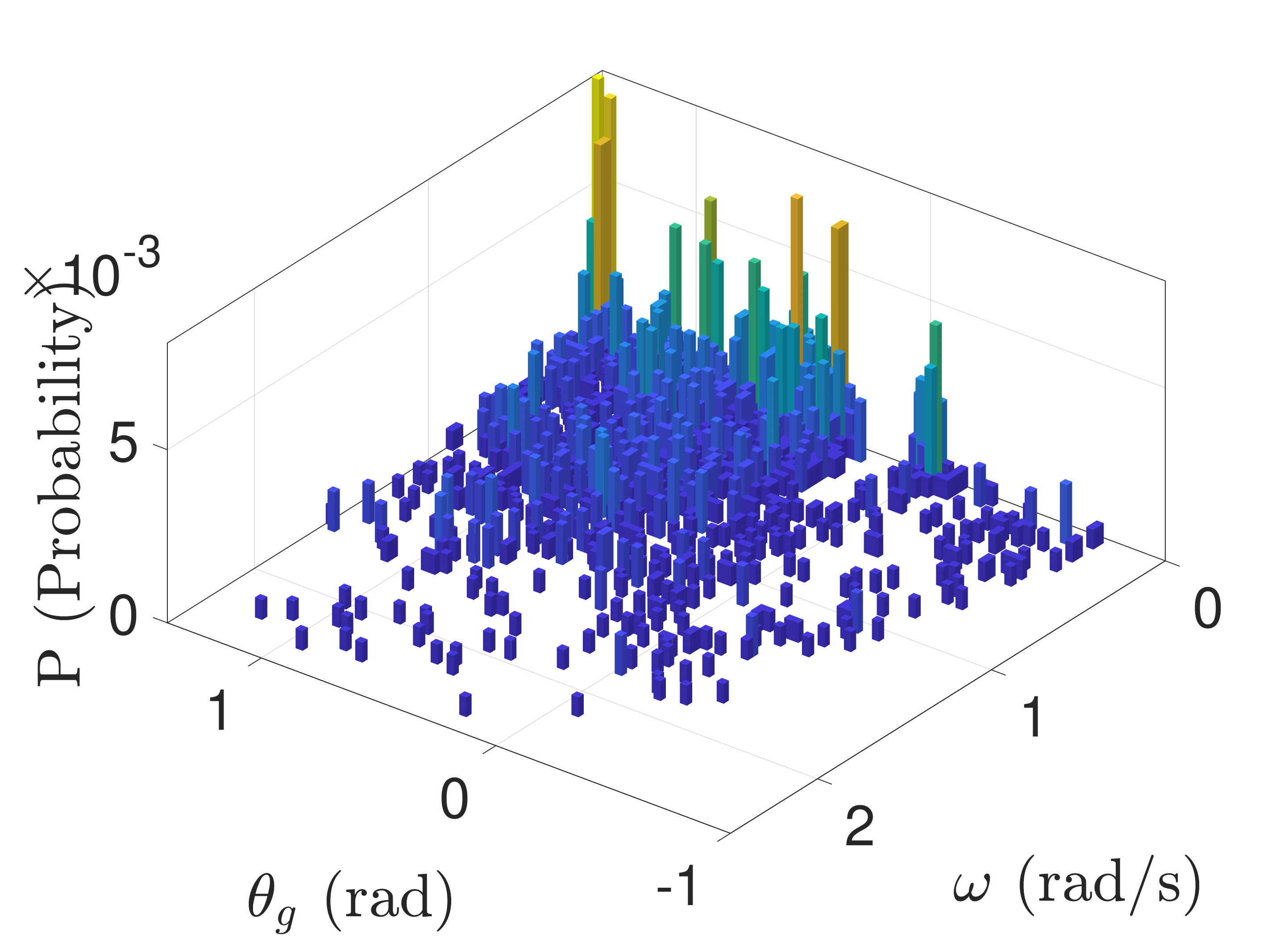}				
	\caption{The probability distribution dataset with recorded participants safe motions.}\label{Fig:Dataset_Prob_C1}
\end{figure}

Finally, in order to find a simple way to evaluate safety $S_f$ at various levels in real-time \footnote{The results of the function are shown in the video.}, we utilize the frequency domain functions as follow
\begin{equation}
	S_f= \begin{cases}
		& \textnormal{Low},  \;\;\;\;\;\;\;\; \;\;\;   \varepsilon_{e,c}<E_{m,s}(f)\\
		& \textnormal{Medium}, \;\;\;\; \varepsilon_{e,m} < E_{m,s}(f)\leq \varepsilon_{e,c} \\
		& \textnormal{High}, \;\;\;\;\;\;\;\;\;\;\;\;\;\;\;\;\;\;\;\;  E_{m,s}(f) \leq \varepsilon_{e,m}
	\end{cases}
\label{Eq:Safetyformulationcase}
\end{equation}
where $\{$low, medium and high$\}$ are the level of safeties, $\varepsilon_{e,m}$ and $\varepsilon_{e,c}$ are the minimum error and maximum critical error ranges, and also $E_{m,q}$ is the mean RMS error in frequency domain (using the Fourier transform) that is found by
\begin{eqnarray}
E_{m,q} &=& \frac{1}{\lambda} \int_{\lambda_m}^{\lambda} E_s(\Omega)  df \nonumber\\
&=&\frac{1}{\lambda} \int_{\lambda_m}^{\lambda} \left( \int_{\infty}^{-\infty} e_q(t) e^{-j\Omega t}  dt \right)  df, 
\label{Eq:thefrequencytransformfourier}
\end{eqnarray}
where $\lambda$ is the frequency range, $\lambda_m$ is the minimum permitted frequency, ${\small e_{q}=(1/n^2) \parallel \bm{\theta} - \bm{\theta}_m \parallel }$ with $t \in [t_0,t_1]$ and $q$ configuration either stands for the angular orientation or velocity, $q=\{\theta,\omega\}$. We have determined the frequency intervals ($\varepsilon_{e,c},\varepsilon_{e,m}$) in  (\ref{Eq:thefrequencytransformfourier}) based on our experiments in multiple people. We think this interval requires more in-depth study since, based on a person's edge, reaction time, and disabilities, it should be chosen. Also, more complex evaluation formulations can be obtained based on human motion patterns by using the obtained real-time physical properties of the safety model ($\bm{\theta},\dot{\bm{\theta}}$).

\section{Results and Discussion}
\label{Sec:ResutlsandDISUCTION}
In this section, we evaluate the developed safety model. At first, we discuss the safety dataset that is prepared. Next, we present our experimental results in a scenario. 

\begin{figure}[t!]
	\centering
	\vspace{3mm} %5mm vertical space	
	\includegraphics[width=2.6 in]{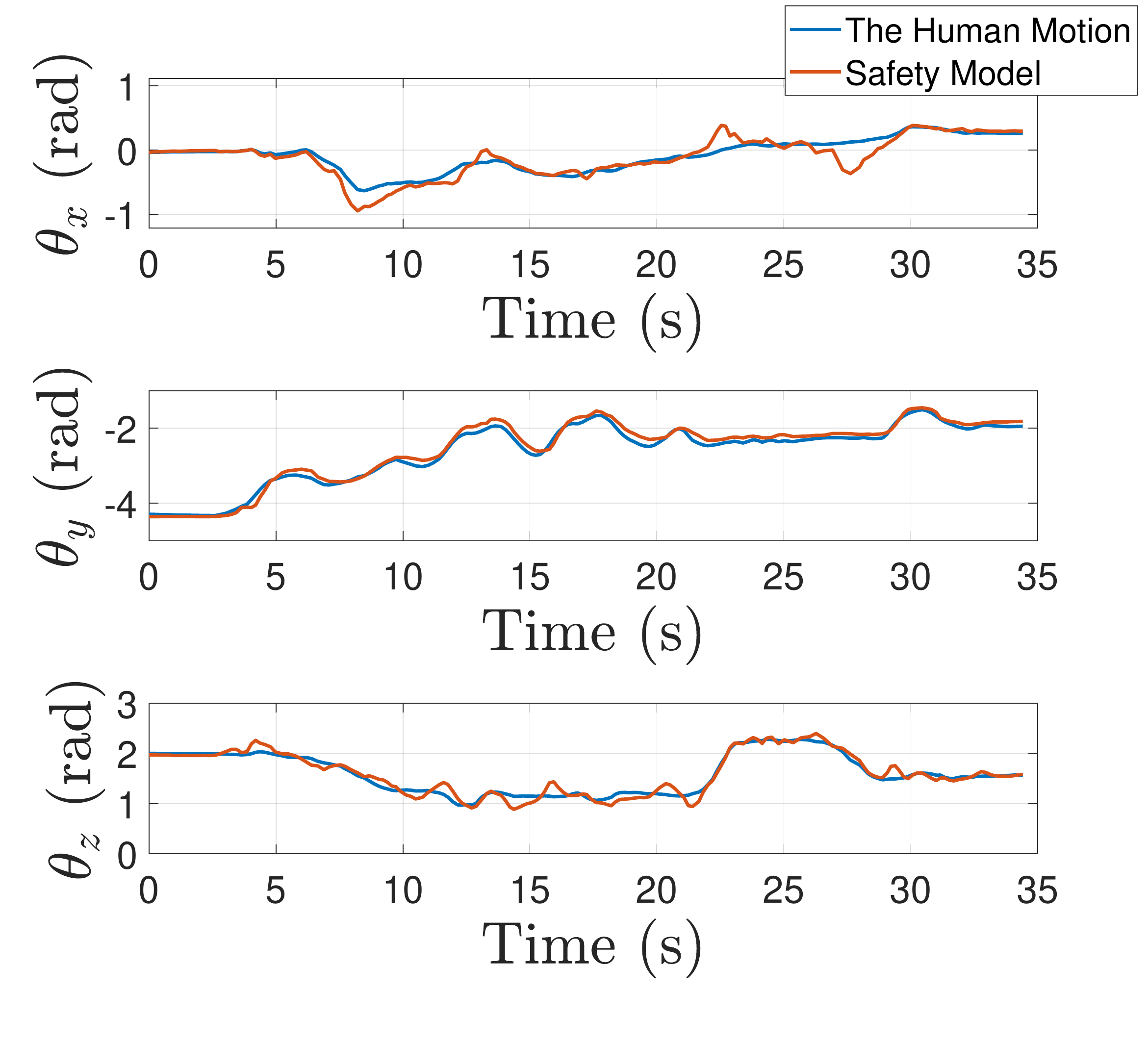}	
	\includegraphics[width=2.6 in]{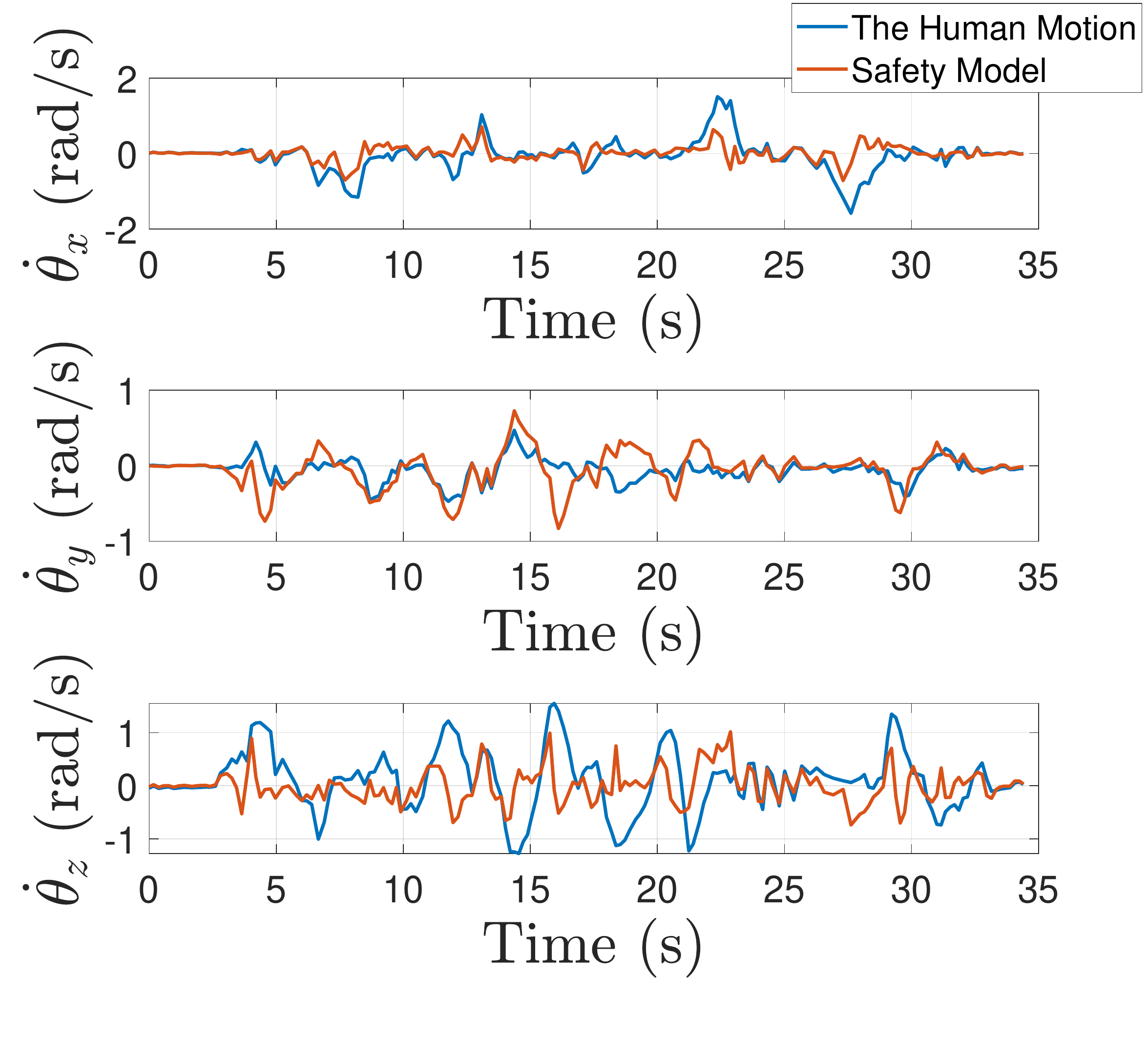}				
	\caption{The first case with normal motions of participant.}\label{Fig:Normal_Motion1}
\end{figure}
\begin{figure}[t!]
	\centering
	\vspace{2mm} %5mm vertical space	
	\includegraphics[width=2.6 in]{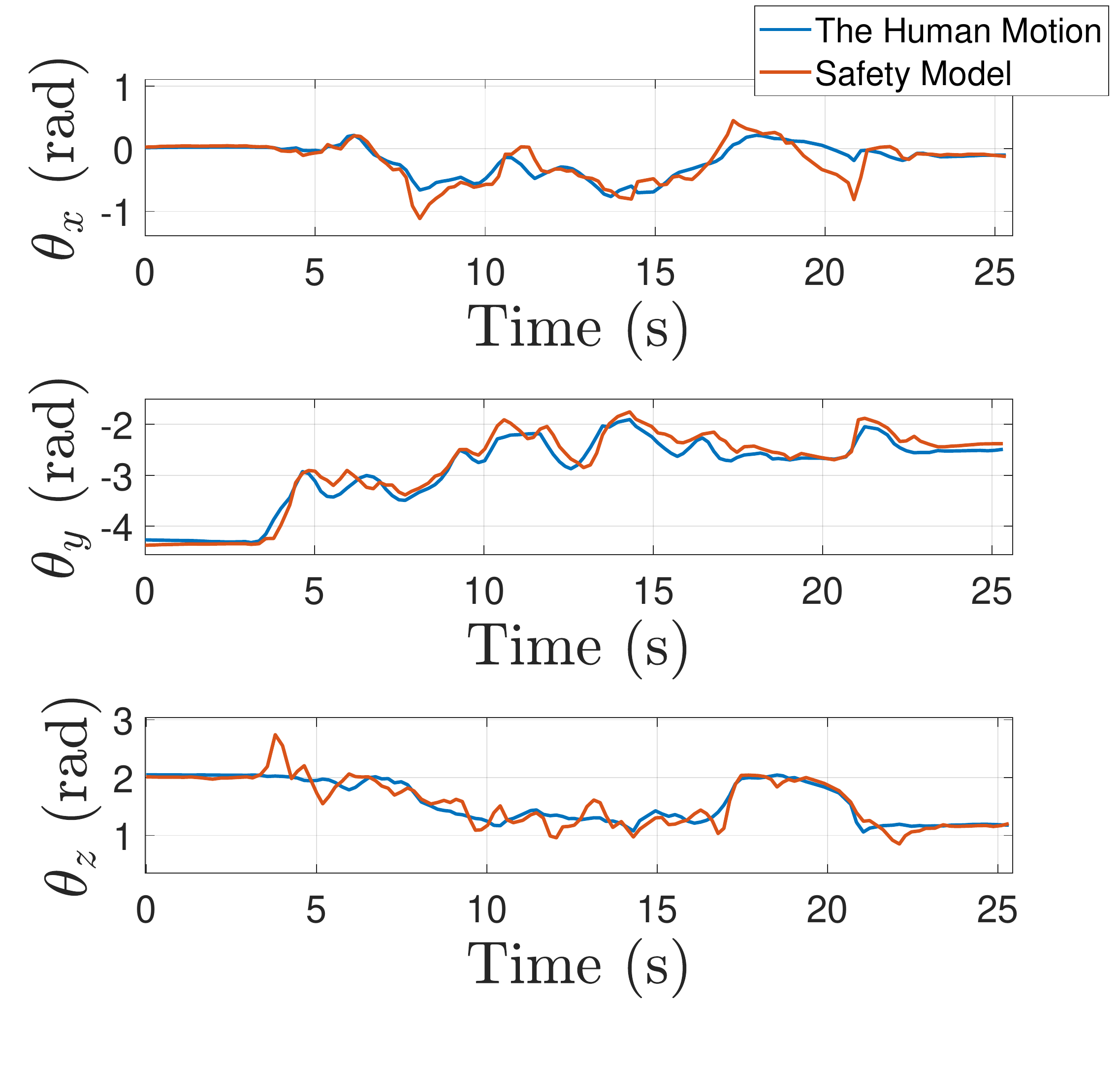}	
	\includegraphics[width=2.6 in]{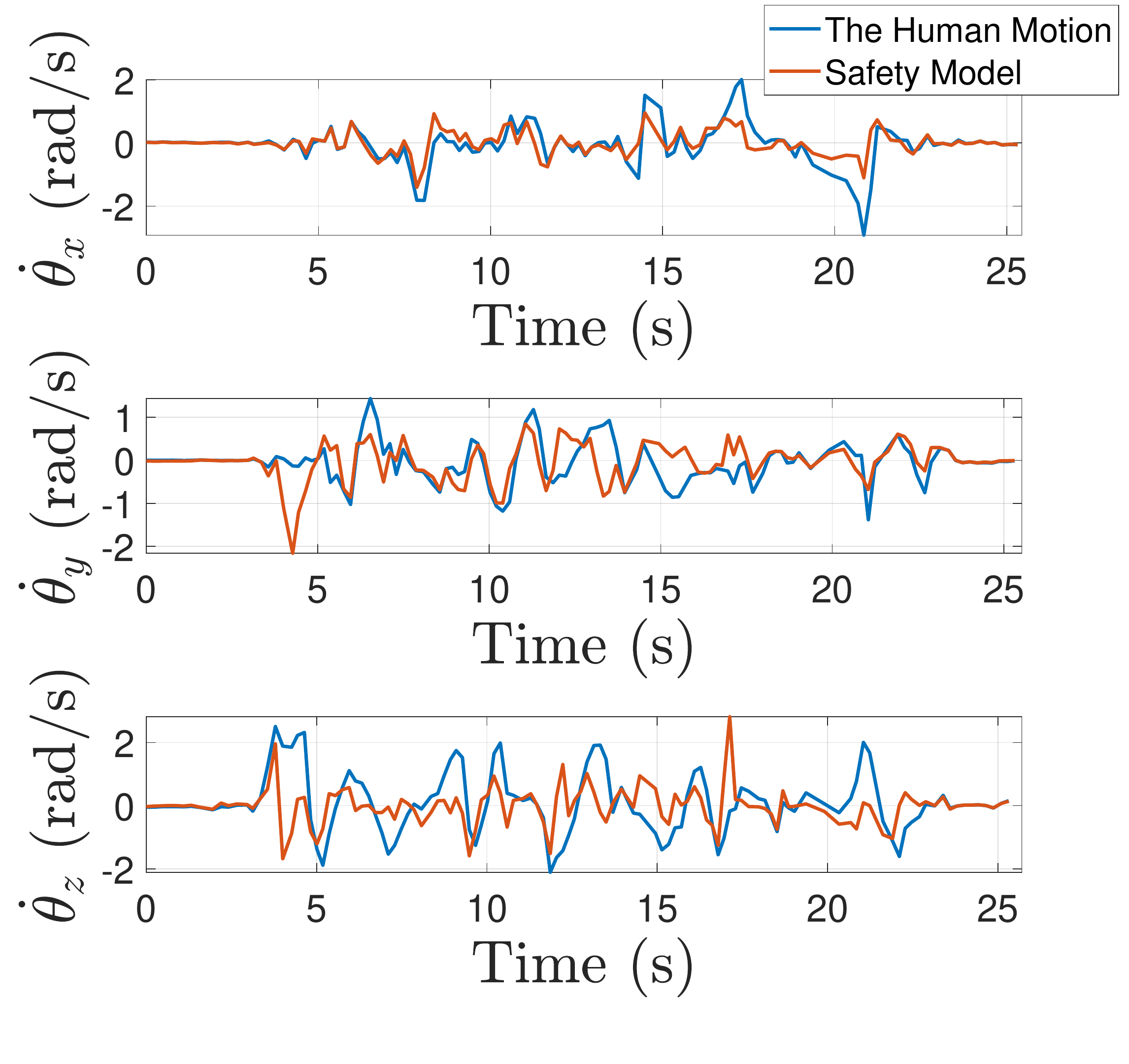}				
	\caption{The second case with unsafe and unstable motions of participant.}\label{Fig:UnNormal_Motion1}
\end{figure}
The dataset is collected based on the proposed structure in Section \ref{Sec:DatasetProblity}. To show that this method can work with a minimum number of recorded data for constructing the probability map, we have used only two participants in this experiment; however, the results are checked with a total of five participants with an average age of 24. The informed consent of participants has been obtained with approval under the Ethics Committee on Research Involving Human Subjects of the Graduate School of Engineering, Tohoku University. The real-world scenario is considered as a person who goes from the bedroom to the washroom as shown in Fig. \ref{Fig:Scenario_Paper_Safety}. To construct the safety dataset, the participants were asked to do a series of discrete motions. These discrete motions were recorded in the following forms: $1\rightarrow2$ (getting up), $\;2\rightarrow3$ (rotating outward on the bed),$\;3\rightarrow4$ (wearing shoes),$\;3\rightarrow5$ (standing up from bed),$\; 7\rightarrow8$ (sitting on the toilet) as shown in the scenario. To keep the data unbiased, the recording happens by creating a standard Gaussian distribution for each part; hence, we ask the participants to move with two slow velocities (safest), two medium velocities, and one high velocity. After processing the collected information, we obtain the presented map in Fig. \ref{Fig:Dataset_Prob_C1} for the probability distribution based on $\left(\theta_g,\omega\right)$. 
\begin{table}[t!]
	\caption{Parameters of the safety model.}
	\label{Tab:Parametervalues}
	\centering
	\begin{tabular}{cccc}
		\hline
		Variable & Value & Variable & Value\\
		\hline
		$m_b$ & 20 kg & $n_f$ & 12 \\
		$l_b$ & 0.2 m& $f_c$ & 0.42 \\
		$\uvec{K}_c$ & $10^2\times[5, 5 ,12]^T$ & $J_{b,s}$ & 0.4\\
		$\uvec{B}_c$  &  $10\times[4, 4 , 6]^T$ &  $\theta_{g,min}$ & $-1$ rad  \\
		$g$ & 9.8  &  $\theta_{g,max}$ & $\pi$/2 rad  \\
		$r_b$ & 0.25 m & $\omega_{max}$& 2.5 rad/s\\
		$\varepsilon_{\theta}$ & 0.03 & $\varepsilon_{e,m}$  &  0.022  \\
		 $\varepsilon_{\omega}$ & 0.03 & $\varepsilon_{e,c}$  & 0.035\\
		
	\end{tabular}
\end{table}

The parameters for the current experiment are given in Table \ref{Tab:Parametervalues}. The physical parameters ($m_b$ and $l_b$) are given for participant shown in Fig. \ref{Fig:Scenario_Paper_Safety}. We experiment the safety model in two cases: first a normal motion and second with unstable and unsafe motion. The human upper body inertia is considered as cylinder geometry with $J_{b,x}=J_{b,y}=(1/12)m_b(3r_b^2+l_b^2)$ and $J_{b,z}=(1/2)m_br_b^2$. The experiment results are shown in the video. In the video, to simply show the different safety levels, we used RMS error of the orientation and angular velocity as derived in (\ref{Eq:Safetyformulationcase}).

\begin{figure}[t!]
	\centering
	\vspace{3mm} %5mm vertical space	
	\includegraphics[width=3.1 in, height=1.3 in]{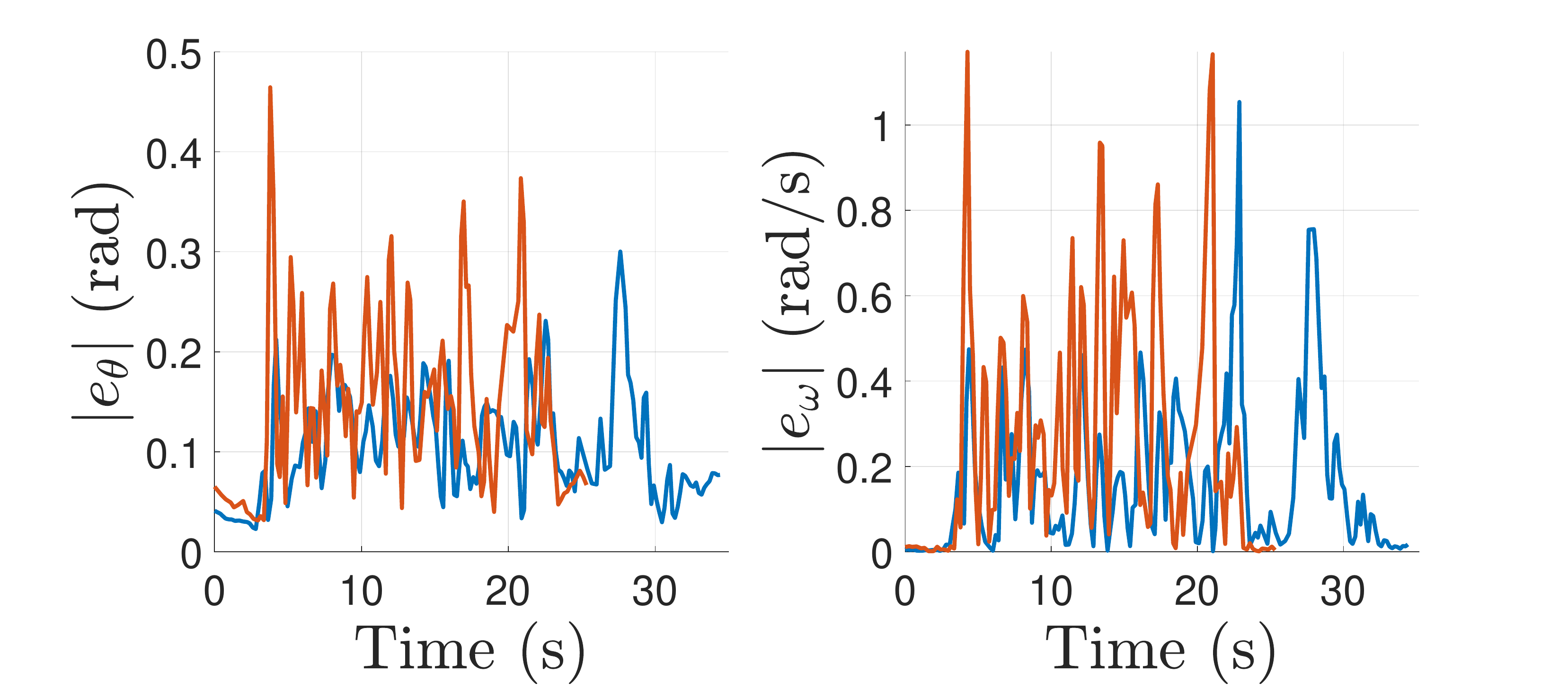}	
	\includegraphics[width=3.2 in, height=1.2 in]{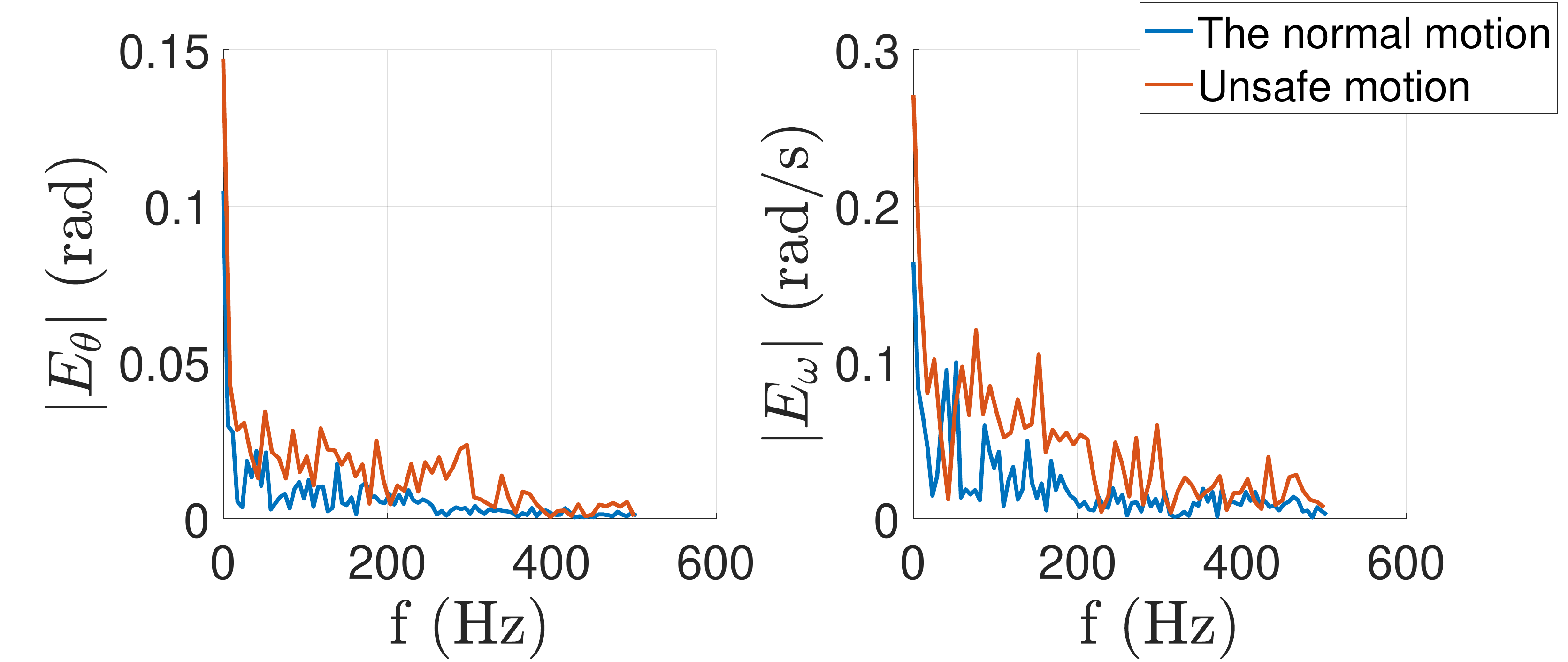}				
	\caption{The RMS error for the angular orientation and velocities during the experiment and the Fourier transform of the whole scenario.}\label{Fig:Compare_Motion1}
\end{figure}
\begin{figure}[t!]
	\centering
	\vspace{3mm}
	\includegraphics[width=3.4 in]{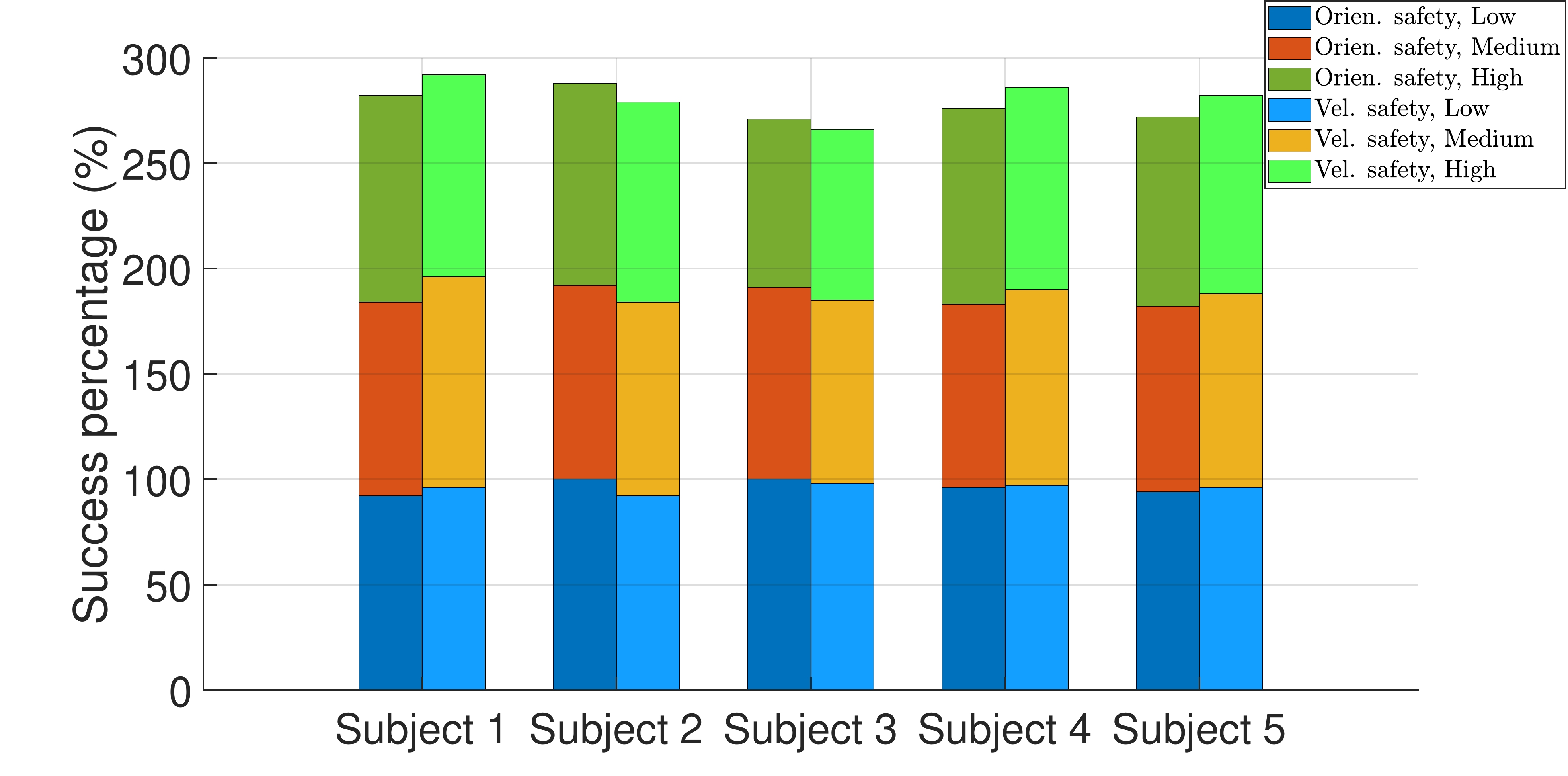}			
	\caption{The success percentage with different participants. Note that the low, medium and high stand for the participant's velocities and medium and mostly high velocities encompass cases with unstable motions.}\label{Fig:SuccessPercentage1}
\end{figure}

Fig. \ref{Fig:Normal_Motion1} shows the first case with the normal motion of the participant. The sensor computation takes between 10-15 Hz frequency. It is clear that the safety model follows human motion, in particular for the angular orientations. This confirms that the human moves safely from the bed to the washroom. In the next case, the participant follows an unstable and continuously changing pattern of motions. The results in both angular and angular velocity differences in Fig. \ref{Fig:UnNormal_Motion1} illustrate that the motions have considerable deviations and this constraint (connected spring-damper pendulum model) shows how much human is not following a proper way orientation. This confirms that our proposed safety model works flawlessly and can determine the unsafe motions in different patterns of the human's upper body motion by certain deviations from the human angular and velocity trajectories.

In order to have a better evaluation of the experiment, we check the RMS error for both angular orientation $|e_\theta|$ and velocity $|e_\omega|$ as shown in Fig. \ref{Fig:Compare_Motion1}. It is clear that during the unsafe motion of the participant, the error is larger than the normal motion case. This issue is much more predominant regarding the velocity demonstrating that the participant has large velocity changes without following normal physical motion. This pattern clarifies that combining a map of the safety dataset with an estimated spring-damper pendulum model gives proper quantification of the human quality of motion. Because the human's random motions are normally inevitable, transforming the RMS error to frequency domain can clearly inform the error repetition. This matter shows the drifts in the amplitude of both high and low-frequency errors (unsafe human motion) in contrast to normal motion. Based on this motivation, we have developed a real-time human motion evaluation with different safety levels that are presented in the video. The angular and velocity constraints based on safety are defined by the range of frequencies, confirming how the predictive safety model detects unsafe motions. 
\begin{table}[t!]
	\vspace{2mm}
	\caption{The mean of success percentage}
	\label{tab:successpercentage}
	\begin{tabular}{cccc}
		\hline
		Safety constraints       & Low speed & Medium speed & High speed\\ \hline
		Orientation  &      96.4 \%        & 91 \%   & 91.4 \% \\ 
		Angular Velocity      & 95.8 \%       & 92.8 \%    & 92.6 \%   \\ 
	\end{tabular}
\end{table}

Fig. \ref{Fig:SuccessPercentage1} illustrates the overall success percentage of the whole experiment. To determine the success, we consider each motion case presented in Fig. \ref{Fig:Scenario_Paper_Safety} and determine in recorded frames whether the safety model properly detects the subject's safe motions. At each speed for the participants, we experimented multiple times. Note that for each participant, we have only updated the physical properties of the model, i.e., $m_b$ and $l_b$, and the dataset kept the same. Although we have been very strict in determining the success percentage for the fault flags of safety, we can see our approach can successfully detect safe human motion from unsafe and unstable ones in different velocities. However, one particular participant, subject 3, had fewer success values (85\%-92\%). After a detailed evaluation, we released that the subject's height and weight were low, and his transitions for the motion cases were so quick that they did not give enough time to the safety model to determine the proper constraints. We think this can be solved in two ways: 1. Specifying the probability dataset and parameters for the subject that will use the approach, 2. Increasing the sampling frequency. Table \ref{tab:successpercentage} presents the summarized average success percentage, which gives over $90$ \% successful detection in all speeds. Also, the safety model can easily determine the person's motion safety at low and medium speeds with over $183$ \% success percentage in total. Important to note that we did not fine-tune the safety model for all the participants, which shows a promising point in the generality of the solution; hence, the mean percentage of success can be higher.

\section{Conclusion}
\label{Sec:ConclusionD}
In this paper, we proposed a new predictive safety model (PSM) that can recognize and evaluate the safety of human motion at different levels using a single inertial measurement unit. The model quantifies the safety in real-time based on angular orientation and velocity of human upper-body motions with respect to designed reference dataset. At first, we derived the dynamic model of the human upper body with a connected 3-Dof spring-damper pendulum model. Next, a map of variables was developed to create a safe probability dataset. Also, we proposed prediction formulas by combining the safe probability dataset and the variables of the spring-damper pendulum model. The performance of the method experimented with participants in a real-world scenario where a person moves from bedroom to washroom. The PSM was able to detect the unsafe and unstable motions of the participant successfully.

The PSM can have applications in different places, e.g., healthcare facilities and homes, where the user uses only an IMU sensor on the chest. In particular, this predictive model can be used on older people or somebody with certain health conditions which is one of our future goals. Not only this model detects anomalies in human motions without any complex algorithm, but also it can quantify the safety physically. Moreover, this quantification is returned as angular orientation and velocity that can provide additional data for other health monitoring systems and robots. Thus, this proposed new model can open interesting new studies. For instance, high-level prediction models can be researched to know human intention. In future, We plan to combine the method with other assistive robots to use the provided information for more agile and ergonomic transportation. 
\section{Acknowledgment}
This work was supported by JSPS KAKENHI grant number JP21K20391 and partially Japan Science and Technology Agency (JST) [Moonshot R$\&$D Program] under Grant JPMJMS2034.

\bibliographystyle{IEEEtran}
\bibliography{MotionDarbWheel}

\begin{thebibliography}{10}
\providecommand{\url}[1]{#1}
\csname url@samestyle\endcsname
\providecommand{\newblock}{\relax}
\providecommand{\bibinfo}[2]{#2}
\providecommand{\BIBentrySTDinterwordspacing}{\spaceskip=0pt\relax}
\providecommand{\BIBentryALTinterwordstretchfactor}{4}
\providecommand{\BIBentryALTinterwordspacing}{\spaceskip=\fontdimen2\font plus
\BIBentryALTinterwordstretchfactor\fontdimen3\font minus
  \fontdimen4\font\relax}
\providecommand{\BIBforeignlanguage}[2]{{%
\expandafter\ifx\csname l@#1\endcsname\relax
\typeout{** WARNING: IEEEtran.bst: No hyphenation pattern has been}%
\typeout{** loaded for the language `#1'. Using the pattern for}%
\typeout{** the default language instead.}%
\else
\language=\csname l@#1\endcsname
\fi
#2}}
\providecommand{\BIBdecl}{\relax}
\BIBdecl

\bibitem{iso2011iso}
I.~O. for Standardization, ``Iso 10218 1 2011 robots and robotic devices safety
  requirements for industrial robots part 1 robots,'' in \emph{International
  Standards Organization Geneva}, 2011.

\bibitem{iso2016ts}
------, ``Ts 15066 2016 robots and robotic devices, collaborative robots,'' in
  \emph{International Standards Organization Geneva}, 2016.

\bibitem{bixby1991proactive}
K.~Bixby, ``Proactive safety design and implementation criteria,'' in
  \emph{Proceedings of National Robot Safety Conference}, 1991.

\bibitem{alami2006safe}
R.~Alami, A.~Albu-Sch{\"a}ffer, A.~Bicchi, R.~Bischoff, R.~Chatila, A.~De~Luca,
  A.~De~Santis, G.~Giralt, J.~Guiochet, G.~Hirzinger \emph{et~al.}, ``Safe and
  dependable physical human-robot interaction in anthropic domains: State of
  the art and challenges,'' in \emph{2006 IEEE/RSJ International Conference on
  Intelligent Robots and Systems}.\hskip 1em plus 0.5em minus 0.4em\relax IEEE,
  2006, pp. 1--16.

\bibitem{colgate2008safety}
E.~Colgate, A.~Bicchi, M.~A. Peshkin, and J.~E. Colgate, ``Safety for physical
  human-robot interaction,'' in \emph{Springer handbook of robotics}.\hskip 1em
  plus 0.5em minus 0.4em\relax Springer, 2008, pp. 1335--1348.

\bibitem{vasic2013safety}
M.~Vasic and A.~Billard, ``Safety issues in human-robot interactions,'' in
  \emph{2013 ieee international conference on robotics and automation}.\hskip
  1em plus 0.5em minus 0.4em\relax IEEE, 2013, pp. 197--204.

\bibitem{ikeura1995variable}
R.~Ikeura and H.~Inooka, ``Variable impedance control of a robot for
  cooperation with a human,'' in \emph{Proceedings of 1995 IEEE International
  Conference on Robotics and Automation}, vol.~3.\hskip 1em plus 0.5em minus
  0.4em\relax IEEE, 1995, pp. 3097--3102.

\bibitem{li2017adaptive}
Z.~Li, J.~Liu, Z.~Huang, Y.~Peng, H.~Pu, and L.~Ding, ``Adaptive impedance
  control of human--robot cooperation using reinforcement learning,''
  \emph{IEEE Transactions on Industrial Electronics}, vol.~64, no.~10, pp.
  8013--8022, 2017.

\bibitem{rojas2019variational}
R.~A. Rojas, M.~A.~R. Garcia, E.~Wehrle, and R.~Vidoni, ``A variational
  approach to minimum-jerk trajectories for psychological safety in
  collaborative assembly stations,'' \emph{IEEE Robotics and Automation
  Letters}, vol.~4, no.~2, pp. 823--829, 2019.

\bibitem{gualtieri2021emerging}
L.~Gualtieri, E.~Rauch, and R.~Vidoni, ``Emerging research fields in safety and
  ergonomics in industrial collaborative robotics: A systematic literature
  review,'' \emph{Robotics and Computer-Integrated Manufacturing}, vol.~67, p.
  101998, 2021.

\bibitem{kadir2021human}
B.~A. Kadir and O.~Broberg, ``Human-centered design of work systems in the
  transition to industry 4.0,'' \emph{Applied Ergonomics}, vol.~92, p. 103334,
  2021.

\bibitem{tseng2021sustainable}
M.-L. Tseng, T.~P.~T. Tran, H.~M. Ha, T.-D. Bui, and M.~K. Lim, ``Sustainable
  industrial and operation engineering trends and challenges toward industry
  4.0: A data driven analysis,'' \emph{Journal of Industrial and Production
  Engineering}, vol.~38, no.~8, pp. 581--598, 2021.

\bibitem{pedrocchi2013safe}
N.~Pedrocchi, F.~Vicentini, M.~Matteo, and L.~M. Tosatti, ``Safe human-robot
  cooperation in an industrial environment,'' \emph{International Journal of
  Advanced Robotic Systems}, vol.~10, no.~1, p.~27, 2013.

\bibitem{zanchettin2015safety}
A.~M. Zanchettin, N.~M. Ceriani, P.~Rocco, H.~Ding, and B.~Matthias, ``Safety
  in human-robot collaborative manufacturing environments: Metrics and
  control,'' \emph{IEEE Transactions on Automation Science and Engineering},
  vol.~13, no.~2, pp. 882--893, 2015.

\bibitem{unhelkar2018human}
V.~V. Unhelkar, P.~A. Lasota, Q.~Tyroller, R.-D. Buhai, L.~Marceau, B.~Deml,
  and J.~A. Shah, ``Human-aware robotic assistant for collaborative assembly:
  Integrating human motion prediction with planning in time,'' \emph{IEEE
  Robotics and Automation Letters}, vol.~3, no.~3, pp. 2394--2401, 2018.

\bibitem{kanazawa2019adaptive}
A.~Kanazawa, J.~Kinugawa, and K.~Kosuge, ``Adaptive motion planning for a
  collaborative robot based on prediction uncertainty to enhance human safety
  and work efficiency,'' \emph{IEEE Transactions on Robotics}, vol.~35, no.~4,
  pp. 817--832, 2019.

\bibitem{chan2020collision}
C.-C. Chan and C.-C. Tsai, ``Collision-free speed alteration strategy for human
  safety in human-robot coexistence environments,'' \emph{IEEE Access}, vol.~8,
  pp. 80\,120--80\,133, 2020.

\bibitem{bajcsy2020robust}
A.~Bajcsy, S.~Bansal, E.~Ratner, C.~J. Tomlin, and A.~D. Dragan, ``A robust
  control framework for human motion prediction,'' \emph{IEEE Robotics and
  Automation Letters}, vol.~6, no.~1, pp. 24--31, 2020.

\bibitem{goutsu2018classification}
Y.~Goutsu, W.~Takano, and Y.~Nakamura, ``Classification of multi-class daily
  human motion using discriminative body parts and sentence descriptions,''
  \emph{International Journal of Computer Vision}, vol. 126, no.~5, pp.
  495--514, 2018.

\bibitem{kratzer2020prediction}
P.~Kratzer, M.~Toussaint, and J.~Mainprice, ``Prediction of human full-body
  movements with motion optimization and recurrent neural networks,'' in
  \emph{2020 IEEE International Conference on Robotics and Automation
  (ICRA)}.\hskip 1em plus 0.5em minus 0.4em\relax IEEE, 2020, pp. 1792--1798.

\bibitem{vianello2021human}
L.~Vianello, J.-B. Mouret, E.~Dalin, A.~Aubry, and S.~Ivaldi, ``Human posture
  prediction during physical human-robot interaction,'' \emph{IEEE Robotics and
  Automation Letters}, vol.~6, no.~3, pp. 6046--6053, 2021.

\bibitem{shafti2019real}
A.~Shafti, A.~Ataka, B.~U. Lazpita, A.~Shiva, H.~A. Wurdemann, and
  K.~Althoefer, ``Real-time robot-assisted ergonomics,'' in \emph{2019
  International Conference on Robotics and Automation (ICRA)}.\hskip 1em plus
  0.5em minus 0.4em\relax IEEE, 2019, pp. 1975--1981.

\bibitem{roetenberg2005compensation}
D.~Roetenberg, H.~J. Luinge, C.~T. Baten, and P.~H. Veltink, ``Compensation of
  magnetic disturbances improves inertial and magnetic sensing of human body
  segment orientation,'' \emph{IEEE Transactions on neural systems and
  rehabilitation engineering}, vol.~13, no.~3, pp. 395--405, 2005.

\bibitem{Tafrishi2021Sen}
S.~A. Tafrishi, M.~Svinin, and M.~Yamamoto, ``A motion estimation filter for
  inertial measurement unit with on-board ferromagnetic materials,'' \emph{IEEE
  Robotics and Automation Letters}, vol.~6, no.~3, pp. 4939--4946, 2021.

\bibitem{mitra2006digital}
S.~K. Mitra and Y.~Kuo, \emph{Digital signal processing: a computer-based
  approach}.\hskip 1em plus 0.5em minus 0.4em\relax McGraw-Hill New York, 2006,
  vol.~2.

\bibitem{gavrila1999visual}
D.~M. Gavrila, ``The visual analysis of human movement: A survey,''
  \emph{Computer vision and image understanding}, vol.~73, no.~1, pp. 82--98,
  1999.

\bibitem{najafi2003ambulatory}
B.~Najafi, K.~Aminian, A.~Paraschiv-Ionescu, F.~Loew, C.~J. Bula, and
  P.~Robert, ``Ambulatory system for human motion analysis using a kinematic
  sensor: monitoring of daily physical activity in the elderly,'' \emph{IEEE
  Transactions on biomedical Engineering}, vol.~50, no.~6, pp. 711--723, 2003.

\bibitem{prakash2018recent}
C.~Prakash, R.~Kumar, and N.~Mittal, ``Recent developments in human gait
  research: parameters, approaches, applications, machine learning techniques,
  datasets and challenges,'' \emph{Artificial Intelligence Review}, vol.~49,
  no.~1, pp. 1--40, 2018.

\end{thebibliography}

\end{document}